\documentclass[10pt,journal]{IEEEtran}
\usepackage[pdftex]{graphicx}
\DeclareGraphicsExtensions{.pdf,.jpeg,.png}

\usepackage{multirow,setspace,verbatim,amsfonts,graphicx,amsmath,amsthm,amsbsy,amssymb,epsfig,url,cite}
\usepackage{booktabs}

\usepackage{amsthm}
\usepackage{gensymb}
\usepackage{graphicx}
\usepackage{amsfonts}
\usepackage{amsmath,bbm}
\usepackage{amssymb}
\usepackage{amsthm}
\usepackage{tikz,graphicx}
\usepackage{mathrsfs}
\usepackage{algpseudocode}

\usepackage{algorithmicx}

\usepackage{array}
\usepackage{booktabs}

\newcommand{\Wb}{{\mathbf W}}

\newcommand{\Yb}{{\mathbf Y}}

\newcommand{\ab}{{\mathbf a}}

\newcommand{\wb}{{\mathbf w}}

\newcommand{\yb}{{\mathbf y}}
\newcommand{\zb}{{\mathbf z}}

\newcommand{\Thetab}{{\boldsymbol {\Theta}}}

\begin{document}

\title{Universal  Deep  Beamformer for Variable Rate  Ultrasound Imaging}

\author{
Shujaat Khan,
        Jaeyoung Huh,~%\IEEEmembership{Student Member,~IEEE,}
%XXX, 
        and~Jong~Chul~Ye,~\IEEEmembership{Senior Member,~IEEE}% <-this % stops a space
%\thanks{Copyright (c) 2017 IEEE. Personal use of this material is permitted. However, permission to use this material for any other purposes must be obtained from the IEEE by sending a request to pubs-permissions@ieee.org.}
\thanks{The authors are with the Department of Bio and Brain Engineering, Korea Advanced Institute of Science and Technology (KAIST), 
		Daejeon 34141, Republic of Korea (e-mail:\{shujaat,woori93,jong.ye\}@kaist.ac.kr). }% <-this % stops a space
}

% make the title area
\maketitle

% As a general rule, do not put math, special symbols or citations in the abstract or keywords.
\begin{abstract}
%In focused B-mode US imaging,  the delay-and-sum (DAS) method is considered as a standard beam-forming method. 
 Ultrasound (US) imaging is based on the time-reversal principle, in which
 individual channel RF measurements are back-propagated and accumulated to form an  image after  applying specific delays. 
While this time reversal is usually implemented as a delay-and-sum (DAS) beamformer,
the image quality quickly degrades as the number of measurement channels decreases.
To address this problem,  various types of adaptive beamforming techniques have been proposed using
predefined models of the signals. Unfortunately, the performance of
these adaptive beamforming approaches degrade when the underlying model is not sufficiently accurate.
Here, we demonstrate for the first time that a single  {\em universal deep beamformer} trained using a purely data-driven way
 can generate 
significantly improved  images  over widely varying aperture and channel subsampling patterns. % range of data subsampling.
In particular, 
 we design an end-to-end deep learning framework that can directly process sub-sampled RF data acquired at different subsampling
 rate and detector configuration to generate high quality ultrasound images  using a single beamformer.
 Experimental results using B-mode focused ultrasound  confirm the efficacy of the proposed
 methods.
\end{abstract}

% Note that keywords are not normally used for peerreview papers.

\IEEEpeerreviewmaketitle

\section{Introduction}
\label{sec:introduction}

Excellent temporal resolution with reasonable image quality makes ultrasound (US) modality a first choice for variety of clinical applications . Moreover, due to its  minimal invasiveness from non-ionizing radiations, 
US  is an indispensable tool for  some clinical applications such as cardiac,  fetal imaging, etc. % the x-ray CT and MRI imaging is difficult.

The basic imaging principle of US imaging is based on the time-reversal  \cite{TimeReversal1,TimeReversal2}, 
which is based on a mathematical observation
that the  wave operator is self-adjoint.
%and that the corresponding Green's function possesses the reciprocity property.
In other words, the wave operator is invariant under time transformation $t\rightarrow -t$,
and the position of the sources and receivers can be swapped. Therefore, it is
possible to reverse a wave from the measurement positions and different
control times to the source locations and the initial time. Practically, this is done by back-propagating the measured data,  after the delay transformation $t\rightarrow t_{\max}-t$,  through adjoint wave and adding all the contributions.

For example, in focused B-mode US imaging, the return echoes from individual scan line are recorded by the receiver channels,
after which  delay-and-sum (DAS) beamformer applies the time-reversal delay  to the channel measurement and additively
combines them for each time point 
to form images at each scan line.

Despite the simplicity,   large number of receiver elements are
often necessary in time reversal imaging to improve the image quality by reducing the side lobes. Similarly, 
 high-speed analog-to-digital converters (ADC) should be used.  
This is because
 the mathematical theory of time reversal is derived assuming that
  the distance between consecutive receivers is taken to be less than half of the wavelength and the temporal scanning is done at a fine rate so that the relative difference between consecutive scanning times is very small \cite{TimeReversal1,TimeReversal2}.
Therefore, with the limited number of receive channels and ADC resolution,
 DAS beamformer suffers from reduced image resolution
and contrast. 

To address this problem, various adaptive beamforming techniques have been developed over the several decades \cite{AdaptiveBF1,CaponBF,CaponBF2,MVBF1,MVBF2,BSBF1,FastRobustBF,MultiBeamBF1,IterativeBF1}.
The main idea of adaptive beamforming is to change the receive aperture weights based on the received data statistics to improve the resolution
and  enhance the contrast.
 For example, one of the most extensively studied adaptive beamforming technique is the Capon beamforming, also known
 as the minimum variance (MV) beamforming \cite{CaponBF,CaponBF2,MVBF1}. The aperture weight of Capon beamfomer is
 derived by minimizing the side lobe while maintaining the gain at the look-ahead direction.
 Unfortunately,  Capon beamforming is  computational heavy for practical use due to
  the calculation of the covariance matrix and its inverse \cite{MVBF2}. Moreover,
  the performance of Capon beamformer is dependent upon the accuracy of the covariance matrix estimate.
To reduce the complexity, many improved version of MV beamformers have been proposed \cite{MVBF1,MVBF2,BSBF1,FastRobustBF}.
Some of the notable examples includes the beamspace adaptive beamformer \cite{BSBF1},
multi-beam Capon based on multibeam covariance matrices\cite{MultiBeamBF1}.
To improve the robustness of Capon beamformer,  parametric form of
the covariance matrix calculation with iterative update was also proposed  rather than calculating the empirical covariance matrix \cite{IterativeBF1}.

However, Capon beamformer and its variants are usually designed for uniform array, so it is difficult to use for 
the subsampled sparse array that is often used to reduce the power consumption and data rate \cite{schretter2018ultrasound,yoon2018efficient}. 
To address this,  compressed sensing (CS) approaches have been recently studied. In \cite{schretter2018ultrasound}, Colas \textit{et al.} proposed a point-spread-functions based sensing matrix for CS reconstruction. However, the accurate measurement of the
spatially varying point spread function is difficult, which limits the resolution for in vivo experiments \cite{schretter2018ultrasound}.
 In \cite{tur2011innovation,wagner2012compressed,chernyakova2014fourier},  compressive beamforming methods
 were proposed.
  But these approaches usually require changes of ADC part of hardwares.

Recently, inspired by the tremendous success of deep learning,
many researchers have investigated deep learning approaches for various inverse problems \cite{kang2017deep,kang2018deep,chen2017lowBOE,adler2018learned,wolterink2017generative,jin2017deep,han2017framing,wang2016accelerating,hammernik2018learning,schlemper2018deep,zhu2018image,lee2018deep}.
In US literature, the works in \cite{Allman_reviewer,luchies2018deep} were among the first to apply deep learning approaches to US image reconstruction.  In particular, Allman \textit{et al} \cite{Allman_reviewer} proposed a machine learning method to identify and remove reflection artifacts in photo-acoustic channel data.  Luchies and Byram \cite{luchies2018deep} proposed a frequency domain deep learning method
for suppressing off-axis scattering in
ultrasound channel data.
%raw data. 
 In \cite{feigin2018deep}, a deep neural network is designed to estimate the attenuation characteristics of sound in human body. In \cite{perdios2017deep,zhou2018high}, ultrasound image denoising method is proposed for the B-mode and single angle
 plane wave imaging, respectively.
Rather than using  deep neural network as a post processing method, the authors in
 \cite{yoon2018efficient,gasse2017high,MICCAI1,MICCAI2} use
 deep neural networks for the reconstruction of high-quality US images from limited number of received RF data.   
For example, the work in \cite{gasse2017high} uses deep neural network for coherent compound imaging
from small number of plane wave illumination. % related to plane-wave imaging modality. 
In focused B-mode ultrasound imaging, \cite{yoon2018efficient} employs the deep neural network
to interpolate the missing RF-channel data with multiline aquisition for accelerated scanning.
%In \cite{antholzer2018photoacoustic}, deep learning is used for photo acoustic image reconstruction.
In \cite{MICCAI1,MICCAI2}, the authors employ deep neural networks
 for the correction of blocking artifacts in multiline acquisition and transmission scheme.

While these recent deep neural network approaches provide impressive reconstruction
performance,  the current design is not universal in the sense that  the designed
neural network cannot completely replace a DAS beamformer, since they are designed
and trained for specific acquisition scenario.
Similar limitation exists in the classical MV beamformer, since the covariance matrix
is determined by the specific detector geometry, which is difficult to adapt, for example,  to dynamically varying
sparse array \cite{RobustCaponBF}.

Therefore,  one of the most important contributions of this paper is to demonstrate that
 a {\em single}  beamformer can generate high quality images robustly for
 various detector channel configurations and subsampling rates. 
  The main innovation of our {\em universal} deep beamformer 
 comes from one of the most exciting properties of deep neural network - exponentially
 increasing expressiveness \cite{rolnick2017power,telgarsky2016colt,arora2016understanding}.
% 
%In particular, expressiveness of neural networks have been one of the most extensively studied research topics.
%For example,
% Rolnick et al \cite{rolnick2017power} showed that 
%the total number of neurons required to approximate natural
%classes of multivariate polynomials of $n$ variables grows only linearly with $n$
%for deep neural networks, but grows exponentially when merely a single hidden
%layer is allowed.
%%Similar results  have been shown by many other authors \cite{?}.
%%Hansen et al \cite{?}, Telgarsky et al \cite{?}.
For example, 
%Telgarsky constructs
%interesting  classes of  functions that can be only computed efficiently
%by deep ReLU nets, but not by shallower
%networks with a simlar number of parameters \cite{telgarsky2016colt}.
Arora et al \cite{arora2016understanding} showed that
for every natural number $k$ there exists a ReLU network with $k^2$ hidden layers and total size of
$k^2$, which can be represented by $\frac{1}{2}k^{k+1}-1$ neurons with at most $k$-hidden layers. 
All these results agree that the expressive power of deep neural networks increases exponentially with the network depth.
Thanks to the exponential large expressiveness with respect to depth, our novel deep neural network beamformer  can learn
 the mapping to images from various sub-sampled RF measurements,
 and exhibits superior image quality for all sub-sampling rate.
 Another amazing feature of the proposed network is that even though
 the network is trained to learn the mapping from the sub-sampled channel data to
 the B-mode images from full rate DAS images,  the trained neural
 network can utilize the fully sampled RF data furthermore to 
 improve the image contrast even for the full rate cases. 
% The origin of this {\em super-resolution} effects of neural
% network will be also discussed in depth.

This paper is  organized as follows. In Section~\ref{sec:theory},  a brief survey of the existing adaptive beamforming methods are provided, which is followed by
the detailed explanation of the proposed universal beamformer.
Section~\ref{sec:methods} then describes the data set and experimental setup.
Experimental results are provided in Section~\ref{sec:results}, which is followed
by Discussion and Conclusions in Section~\ref{sec:discussion} and Section~\ref{sec:conclusion}.

%Our theoretical understanding of the link led us to implement a neural network fo in the RF to IQ domain, which may be unexpected from the perspective of implementing a conventional deep neural network.  In contrast to the RF to RF domain CNN that attempt to generate reconstructed plane for a specific down-sampling ratio, one of the most important advantages of the this work is that our single CNN-based model can be used for all down-sampling ratios. This shows its marvellous generalization power. 

\section{Theory}
\label{sec:theory}

\subsection{Adaptive Beamforming}

The standard non-gain compensated delay-and-sum (DAS) beamformer for the $l$-th scanline at the depth sample $n$ can be expressed as
\begin{equation}
{z}_l[n] =\frac{1}{J} \sum_{j=0}^{J-1}x_{l,j}[n-\tau_{j}[n]]=\frac{1}{J}\mathbf{1}^\top\mathbf{y}_l[n],  \quad l=0,\cdots, L-1
\end{equation}
where $^\top$ denotes the transpose, $x_{l,j}[n]$ is the  RF echo signal measured by the $j$-th active receiver element from the transmit event (TE) for the
$l$-th scan line,
and $J$ denotes the number of active receivers, 
  $\tau_{j}[n]$ is the dynamic focusing delay for the
 $j$-th active receiver elements to obtain the $l$-th  scan line.
 Furthermore, $\yb_l[n]$  refers to the scan line dependent time reversed RF data defined by
 \begin{eqnarray}
 \mathbf{y}_l[n]=\begin{bmatrix} y_{l,0}[n] & y_{l,1}[n] & \cdots &y_{l,J-1}[n]]\end{bmatrix}^\top
 \end{eqnarray}
 with $y_{l,j}[n]:=x_{l,j}\left[n-\tau_{j}[n]\right] $, and 
 $\mathbf{1}$ denotes a length $J$ column-vector of ones.

This averaging of the time-delayed element-outputs extracts the (spatially) low-frequency content that corresponds to the energy within one scan resolution cell (or main lobe). Reduced side lobe leakage at the expense of a wider resolution cell can be achieved by replacing the uniform weights by tapered weights $w_{l,j}[n]$:
\begin{equation}\label{BF2}
z_l[n] %= \sum_{j=0}^{J-1}w_{j}x_{j}[n-\tau_{j}[n]]
=\sum_{j=0}^{J-1}w_{l,j}[n]y_{l,j}[n] = \wb_l[n]^\top \yb_l[n]
\end{equation}
where
$\mathbf{w}_l[n] =\begin{bmatrix}w_{l,0}[n] &  w_{l,1}[n] & \cdots & w_{l,J}[n]\end{bmatrix}^{\top}.$
In adaptive beamforming the objective is to find the $\mathbf{w}_l$ that minimizes the variance of ${z}_l$, subject to the constraint that the gain in the desired beam direction equals unity. The minimum variance (MV) estimation task can be formulated as  \cite{CaponBF,CaponBF2,MVBF1}
\begin{equation*}
\begin{aligned}
& \underset{\mathbf{w}_l[n]}{\text{minimize}}
& & E[|z_l[n]|^2]=\min_{\mathbf{w}[n]}\mathbf{w}_l[n]^{\top}\mathbf{R}_l[n]\mathbf{w}_l[n] \\
& \text{subject to}
& &  \mathbf{w}_l[n]^{H} \ab =1,
\end{aligned}
\end{equation*}
where $E[\cdot]$ is the expectation operator,  and $\mathbf{a}$ is a steering vector, which is composed of ones when the received signal is already temporally aligned, and $\mathbf{R}[n]$ is a spatial covariance matrix expressed as
\begin{equation}
\mathbf{R}_l[n]=E\left[\mathbf{y}_l[n]^{\top} \mathbf{y}_l[n]\right],
\end{equation}
Then, $\wb_l[n]$ can be obtained by  Lagrange multiplier method \cite{LagrangeBF} and expressed as
\begin{equation}
\mathbf{w}_l[n]=\frac{\mathbf{R}_l[n]^{-1} \mathbf{a}}{\mathbf{a}^{H} \mathbf{R}_l[n]^{-1} \mathbf{a}} .
\end{equation}
In practice, $\mathbf{R}_l[n]$ must be estimated with a limited amount of data. A widely used method for the  estimation of $\mathbf{R}_l[n]$ is spatial smoothing (or subaperture averaging) \cite{AdaptiveBFMUS}, in which  the sample covariance matrix is calcualted  by averaging covariance matrices of $K$ consecutive channels in the $J$ receiving channels as follows:
\begin{equation} \label{BF5}
\mathbf{\tilde{R}}_l[n]=\frac{1}{J-K+1} {\mathbf{Y}_l}[n]{\mathbf{Y}_l}^{\top}[n],
\end{equation}
where
\begin{equation}\label{BF6}
{\mathbf{Y}_l}[n] = \begin{bmatrix} 
y_{l,0}[n] & \dots & y_{l,J-K}[n] \\
%y_{l,1}[n]& \dots & y_{l,J-K+1}[n]\\
\vdots &        & \vdots \\
y_{l,K-1}[n] & \dots & y_{l,J-1}[n] 
\end{bmatrix},
\end{equation}
which is invertible if $K \leq J/2$. 
To further improve the invertibility of the sample covariance matrix,
 another method usually called diagonal loading is often used by adding
additional diagonal terms \cite{AdaptiveBFMUS}.

One of the problems with the MV beamforming technique employed in a medical ultrasound imaging system is that the speckle characteristics tend to be different from those of conventional DAS beamformed images. MV beamformed images tend to look slightly different from conventional DAS B-mode images in that the speckle region appears to have many small black dots interspersed (see \cite[Fig. 6]{AdaptiveBFMUS}). To overcome this problem, a temporal averaging method \cite{BenefitsMV} that averages $\mathbf{\tilde{R}}$ along the depth direction is used, which is expressed as
\begin{equation}\label{BF10}
\mathbf{\tilde{R}}_l[n]=\frac{1}{2L+1} \frac{1}{J-K+1} \sum_{l=-L}^{L}{\mathbf{Y}_l}[n+l]{\mathbf{Y}_l}^{\top}[n+l],
\end{equation}
%
%This $\mathbf{\tilde{R}}[n]$ replaces that in \eqref{BF5} when spatial and temporal averaging are used together. Hereafter, the MV method just described will be called standard MV. We will use only spatial smoothing and diagonal loading throughout this paper, i.e., $L = 0$ in \eqref{BF10}.
%

Another method to estimate the covariance matrix in MV is so-called  multibeam approach \cite{MultiBeamBF1}. In this method,
 the weight vector is estimated using 
%\begin{equation}
%\mathbf{w}[n]=\frac{\mathbf{R_{\Theta}}[n]^{-1} \mathbf{a_{\theta}}}{\mathbf{a_{\theta}}^{H} \mathbf{R_{\Theta}}[n]^{-1} \mathbf{a_{\theta}}} .
%\end{equation}
empirical covariance matrices 
%$\mathbf{R_{\Theta}}$ is a matrix 
that are formed to use phase-based (narrowband) steering vectors  to extract the adaptive array weights from it.

\begin{figure}[!hbt]
   \centering
   \centerline{\epsfig{figure=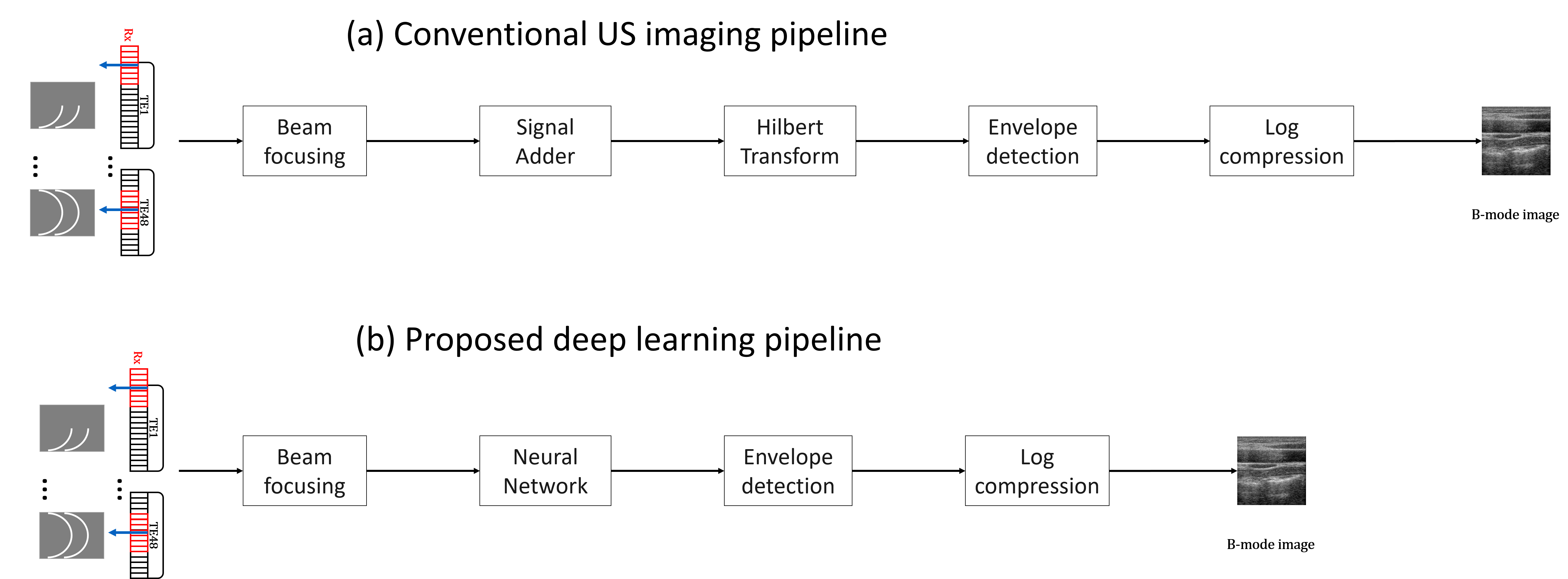, width=9cm,height=5.5cm}}   
    \vspace*{0.3cm} 
    \caption{\footnotesize Ultrasound imaging pipeline. ({a}) standard focused B-mode pipeline, and ({b}) the proposed neural network based reconstruction pipeline.}
    \label{fig:system_block_diagram}
\end{figure}

\begin{figure*}[!hbt]
%   \centering
%   \centerline{\epsfig{figure=CNN_concept_diagram.png, width=18cm}}
%   \vspace*{0.3cm}
%   \centerline{\mbox{(a) Proposed ultrasound imaging system}}
   \centerline{\epsfig{figure=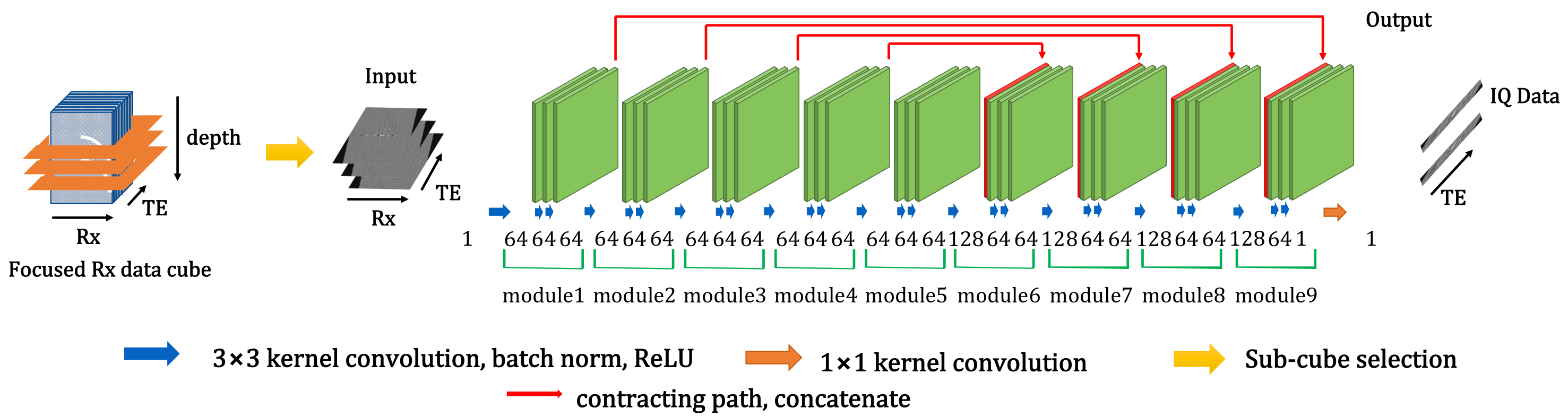, width=18cm}}
   \vspace*{0.3cm}   
%   \centerline{\mbox{(b) Proposed CNN architecture}}
   \caption{\footnotesize Proposed CNN based Ultrasound imaging system block diagram.}
   \label{fig:CNN_block_diagram}
\end{figure*}

\subsection{Proposed algorithm}

\subsubsection{Image reconstruction pipeline}
 Fig.~\ref{fig:system_block_diagram}(a) illustrates the conventional US image reconstruction piplelne.
 Here,  the reflected sound waves in the medium are detected by the transducer elements. Each measured signal is time reversed based on the traveled distance to perform beam-focusing. 
 The focused signals are later added. In this paper, the adaptive beamformer can be used for providing adaptive summation of the time-reversed echos.
 This is then followed by the Hilbert transform  to detector the envelope of the beam.
 In particular, the envelop is determined by calculating the absolute value of the inphase and quadrature pahse signals generated from  the Hilbert transform. % as show in Fig~\ref{fig:system_block_diagram}(a).
 Finally, the log compression is applied to generate the B-mode images.

On the other hand, our goal is to replace the 
the signal adder and Hilbert transformation step by a convolutional neural network (CNN) as shown in Fig.~\ref{fig:system_block_diagram}(b).
Time reversal part is still based on the physical delay calculation, since this is the main idea of the time reversal algorithms.
Envelop detection and log compression are just a simple point-wise operation, so the neural network is not necessary.
Therefore,  our goal is to basically replace the core beamformer and reconstruction engine with a data-driven way CNN.

\subsubsection{Universal Deep Beamformer}

Recall that the basic idea of adaptive beamformer is to estimate the array weight $\wb_l[n]$ from the data to estimate $z_l[n]$, which changes with respect to the scan line index $l$ and the depth $n$.
%IN varies with respect to the scan line index $l$ and the depth $n$.
In the conventional adaptive beamformer, this estimation is usually done based on the linear weight model calculated from the empirical covariance.
However, this  linear model is usually based on restricted assumption, such as  zero mean, Gaussian noise, etc, which may limit
the fundamental performance of the adaptive beamformer.
Moreover, nonlinear beamforming methods have been recently proposed to overcome the limitation of linear model\cite{matrone2015delay,matrone2017high,matrone2017depth,cohen2018sparse}.
Another important step after the beamforming is the Hilbert transform to obtain analytic representation.
More specifically,  Hilbert transform gives the analytic representation of a signal u(t):
\begin{eqnarray}
z_l^a[n]= z_l[n]+ \iota H(z_l) [n] %
\end{eqnarray}% } u_{a}(t)=u(t)+i\cdot H(u)(t)
where $\iota = \sqrt{-1}$, and $H$ denotes the Hibert transform. $z_l^a[n]$ is often referred to as the inphase (I) and quadrature (Q) representation.
To implement Hilbert transform, discrete convolution operation is usually 
performed for each scan line along the depth direction.

One of the main key ideas of the proposed method is a direct
estimation of the beamfored and Hilbert transformed signal $z_l^a[n]$  directly from the time-reverse signal $\yb_l[n]$ using convolutional 
neural network.
To exploit the redundancies along the scan line direction, rather than estimating the beamformed
signal for each scan line, we are interested in estimating the beamformed  and Hilbert transformed signal at whole scan line, i.e.
$$\zb^a[n]= \begin{bmatrix} z_0^a[n] & \cdots & z_{L-1}^a[n]\end{bmatrix}^\top$$
Furthermore, to deal with  the potential blurring along the depth,  we are interested in exploiting the
time reversed signal at three depth coordinates, i.e.
\begin{eqnarray}\label{eq:delay}
\Yb[n] = \begin{bmatrix}  \yb_0[n-1] & \yb_1[n-1] & \cdots & \yb_{L-1}[n-1] \\  \yb_0[n] & \yb_1[n] & \cdots & \yb_{L-1}[n] \\   \yb_0[n+1] & \yb_1[n+1] & \cdots & \yb_{L-1}[n+1]  \end{bmatrix}
\end{eqnarray}
Then, our goal is to estimate the nonlinear function $f(\Wb,\Yb[n])$ such that
$$\zb^a[n] = f(\Thetab,\Yb[n])$$
where $\Thetab$ denotes the trainable CNN parameters.
To generate the complex output,  our neural network generates the two channel output that corresponds to the real
and image parts.
Then,  our CNN called deep beamformer (DeepBP) is trained as follows:
\begin{eqnarray}\label{eq:training}
\min_\Theta  \sum_{i=1}^T \sum_n \|\zb^{a(i)}[n]-f(\Thetab,\Yb^{(i)}[n])\|^2
\end{eqnarray}
where $\zb^{a(i)}[n]$ denotes the ground-truth I-Q channel data at the depth $n$ from the $i$-th training data,
and $\Yb^{(i)}[n]$ represented the corresponding (sub-sampled) time-delayed input data formed by \eqref{eq:delay}.

Note that our current training scheme is depth-independent so that the same CNN can be used across all depth.
Furthermore, as for the target data for the training, we use the standard DAS beamformed data from full detector samples.
Since the target data is obtained from various depth across multiple scan lines, our neural
network is expected to learn the best parameters on averages.
Interestingly, this average behavior turns out to improve the overall image quality even without any subsampling thanks to the
synergistic learning from many training data, as will be shown later in experiments.

Fig.~\ref{fig:CNN_block_diagram} illustrates the schematic diagram of our deep beamformer.
%The proposed method exploits the spatio-temporal redundancy in Rx-TE plane for multiple depths to estimate the weights of the neural network. The idea behind using multiple Rx-TE planes i.e., $\mathbf{\Psi}[n-1]$, $\mathbf{\Psi}[n]$, and $\mathbf{\Psi}[n+1]$ to estimate single $\tilde{\mathbf{Z}}[n]$ is motivated by spatial smoothing \cite{BenefitsMV} and multi-beam covariance estimation method \cite{MultiBeamBF1}. 
In particular, we trained our model with the input/output pairs of a three depth of Rx-TE-Depth data cube as an input and the I-Q data on a single
Rx-TE plane as a target.

\section{Method}
\label{sec:methods}

\subsection{Data set}

For experimental verification, multiple RF data were acquired with the E-CUBE 12R US system (Alpinion Co., Korea).  For data acquisition, we used a linear array transducer (L3-12H) with a center frequency of $8.48$ MHz.  The configuration of the probe is given in Table \ref{probe_config}.

\begin{table}[!hbt]
	\centering
	\caption{Probe Configuration}
	\label{probe_config}
	\resizebox{0.25\textwidth}{!}{
	\begin{tabular}{c|c}
\hline
		{Parameter} & {Linear Probe} \\ \hline\hline
		Probe Model No.& L3-12H \\
		Carrier wave frequency & 8.48 MHz \\
		Sampling frequency & 40 MHz \\
		No. of probe elements & 192 \\
		No. of Tx elements & 128 \\
		No. of TE events & 96\\
		No. of Rx elements & 64\\
		Elements pitch & 0.2 mm\\
		Elements width & 0.14 mm \\
		Elevating length & 4.5 mm\\  \hline
	\end{tabular}}
\end{table}

Using a linear probe, we acquired RF data from the carotid area from $10$ volunteers.  The \textit{in-vivo} data consists of $40$ temporal frames per subject, providing $400$ sets of Depth-Rx-TE data cube.  The dimension of each Rx-TE plane was $64\times96$.  A set of $30,000$ Rx-TE planes was randomly selected from the $4$ subjects datasets, and data cubes (Rx-TE-depth) are then divided into $25,000$ datasets for training and $5000$ datasets for validation.  The remaining dataset of $360$ frames was used as a test dataset.

In addition, we acquired $188$ frames of RF data from the ATS-539 multipurpose tissue mimicking phantom. This dataset was only used for test purposes and no additional training of CNN was performed on it.  The phantom dataset was used to verify the generalization power of the proposed algorithm.

 \begin{figure*}[!hbt]
	\centerline{\includegraphics[width=16cm,height=5cm]{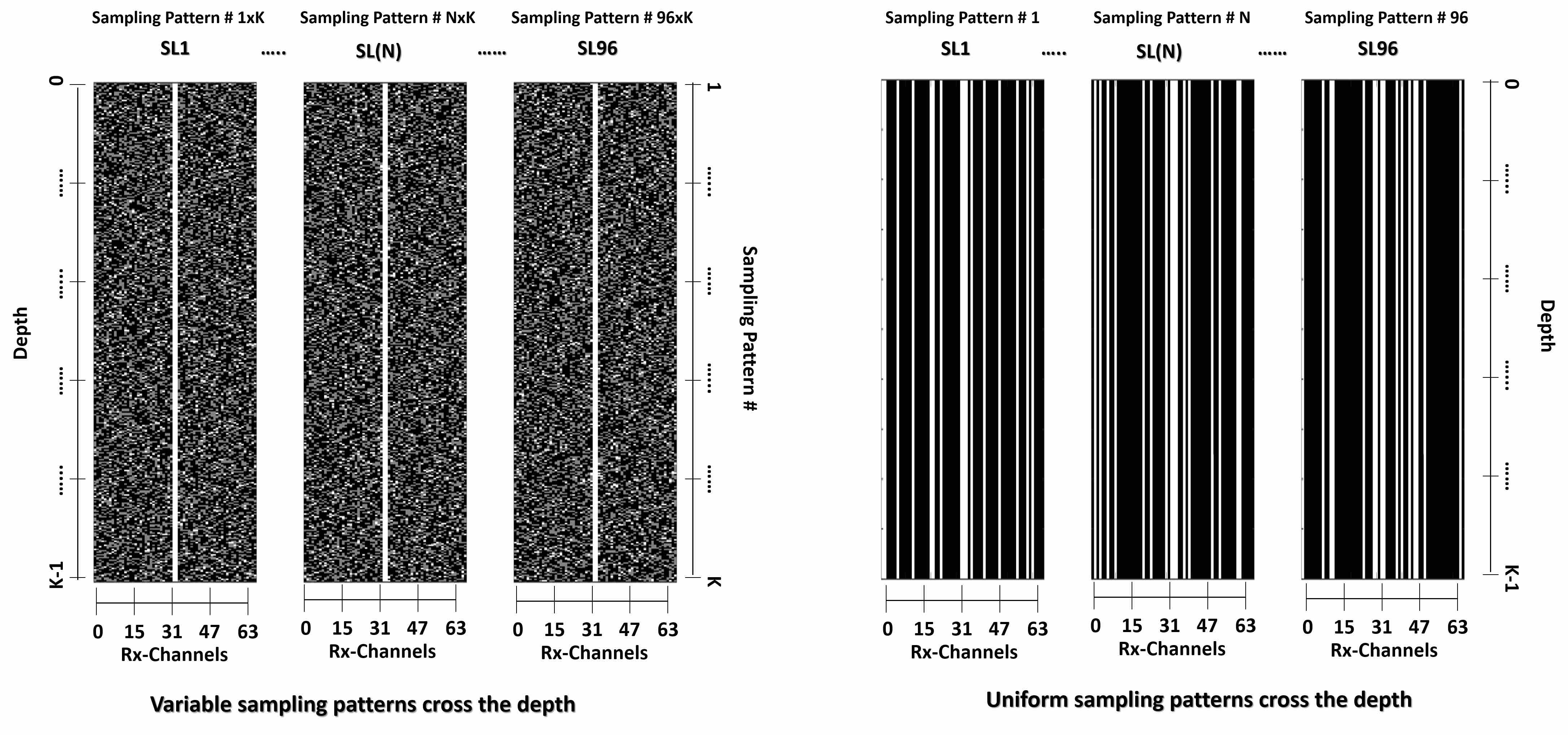}}
%	\vspace*{0.3cm}
	\caption{(left) sampling scheme \# 1: variable sampling cross depth axis. (right) sampling scheme \# 2: uniform sampling cross depth axis.}
	\label{fig:SamplingSchemes}
\end{figure*}

\subsection{RF Sub-sampling Scheme}
\label{sec:RFsamScon}

For our experiments, we generated six sets of sub-sampled RF data at different down-sampling rates. In particular, we use several
subsampling cases using $64$, $32$, $24$, $16$, $8$ and $4$ Rx-channels,  and two subsampling schemes were used: variable down-sampling pattern cross the depth, fixed down-sampling pattern cross the depth (see Figs.~\ref{fig:SamplingSchemes}). 

Since the active receivers at the center of the scan-line get RF data from direct reflection, the two channels that are in the center of active transmitting channels were always included to improve the performance, and remaining channels were randomly selected from the total $64$ active receiving channels. In variable sampling scheme, different sampling pattern (mask) is used for each depth plane whereas, in fixed sampling, we used same sampling pattern (mask) for all depth planes. The network was trained for variable sampling scheme only and both sampling schemes were used in test phase.

\subsection{Network Architecture}
For all sub-sampling schemes samples, a multi-channel CNN was applied to $3\times64\times96$ data-cube in the depth-Rx-TE sub-space to generate a $2\times3\times96$ I and Q data in the depth-TE plane. The target IQ data is obtained from two output channels each representing real and imaginary parts. 

The proposed CNN is composed of convolution layers, batch normalization layers, ReLU layers and a contracting path with concatenation as shown in Figs.~\ref{fig:CNN_block_diagram}(b). Specifically, the network consists of $29$ convolution layers composed of a batch normalization and ReLU except for the last convolution layer. The first $28$ convolution layers use $3\times3$ convolutional filters (i.e. the 2-D filter has a dimension of $3\times 3$), and the last convolution layer uses a $1\times1$ filter and contract the $3\times64\times96$ data-cube from depth-Rx-TE sub-space to $2\times3\times96$ IQ-depth-TE plane.

%The network was implemented with both TensorFlow \cite{abadi2016tensorflow} and MatConvNet \cite{vedaldi2015matconvnet} in the MATLAB 2015b environment to verify the platform-dependent sensitivity.  We found that with the same training strategy, the two implementations provided near identical results. 
The network was implemented with MatConvNet \cite{vedaldi2015matconvnet} in the MATLAB 2015b environment. Specifically, for network training, the parameters were estimated by minimizing the $l_2$ norm loss function. The network was trained using a stochastic gradient descent with a regularization parameter of $10^{-4}$. The learning rate started from $10^{-4}$ and gradually decreased to $10^{-7}$. The weights were initialized using Gaussian random distribution with the Xavier method \cite{glorot2010understanding}. The number of epochs was $200$ for all down-sampling rates.

\subsection{Performance Metrics}

 To quantitatively show the advantages of the proposed deep learning method, we used the contrast-to-noise ratio (CNR) \cite{BiomedicalImageAnalysis}, generalized CNR (GCNR) \cite{GCNR_Paper}, peak-signal-to-noise ratio (PSNR), structure similarity (SSIM) \cite{1284395} and the reconstruction time.  
 
 The CNR is measured for the background ($B$) and anechoic structure ($aS$) in the image, and is quantified as
 \begin{equation}
 {\hbox{CNR}}(B,aS) = \frac{|\mu_{B}-\mu_{aS}|}{\sqrt{\sigma^2_{B} + \sigma^2_{aS}}},
 \end{equation}
  where $\mu_{B}$, $\mu_{aS}$, and $\sigma_{B}$, $\sigma_{aS}$ are the local means, and the standard deviations of the background ($B$) and anechoic structure ($aS$) \cite{BiomedicalImageAnalysis}. 
  
   Recently, an improved measure for the contrast-to-noise-ratio called generalized-CNR (GCNR) is proposed  \cite{GCNR_Paper}. The GCNR compared the overlap between the intensity distributions of two regions.  The GCNR measure is difficult to tweak and shows exact quality improvement for non-linear beam-formers on a fixed scale ranges from zero to one, where one represents no overlap in the distributions of background and region-of-interest (ROI). 
The GCNR is defined as
 \begin{equation}
 {\hbox{GCNR}}(B,aS) = 1- \int \min \{p_{B} (x), p_{aS} (x) \} dx.
 \end{equation}
 where $x$ is the pixel intensity, $p_{B}$ and $p_{aS}$, are the probability distribution of the background ($B$) and anechoic structure ($aS$). If both distribution are completely independent,  then GCNR will be equals to one, whereas, if they completely overlap then GCNR will be zero \cite{GCNR_Paper}.
 
  The PSNR and SSIM index are calculated on reference ($F$) and Rx sub-sampled ($\tilde F$) images of common size  $n_1\times n_2$ as

\begin{equation}
{\hbox{PSNR}}(F,\tilde F) = 10 \ensuremath{\log_{10}} \left(\frac{ n_1n_2 R_{\max}^2}{\|F-\tilde F\|_F^2}\right),
\end{equation}
where $\|\cdot\|_F$  denotes the Frobenius norm and $R_{\max}=2^{(\#bits\ per\ pixel)}-1$ is the dynamic range of pixel values (in our experiments this is equal to $255$),
and
\begin{equation}
{\hbox{SSIM}}(F,\tilde F) = \frac{(2\mu_{F}\mu_{\tilde F}+c_{1})(2\sigma_{F,\tilde F} +c_{2})}{(\mu^{2}_{F}+\mu^{2}_{\tilde F}+c_{1})(\sigma^{2}_{F}+\sigma^{2}_{\tilde F} +c_{2})},
\end{equation}
where  $\mu_{F}$, $\mu_{\tilde F}$, $\sigma_{F}$, $\sigma_{\tilde F}$, and $\sigma_{F,\tilde F}$ are the local means, standard deviations, and cross-covariance for images $F$ and $\tilde F$ calculated for a radius of $50$ units.  The default values of $c_{1}=(k_{1}R_{max})^{2}$, $c_{2}=(k_{2}R_{max})^{2}$, $k_{1}=0.01$ and $k_{1}=0.03$.

\begin{figure*}[!hbt]
%	\centerin
	\centerline{\epsfig{figure=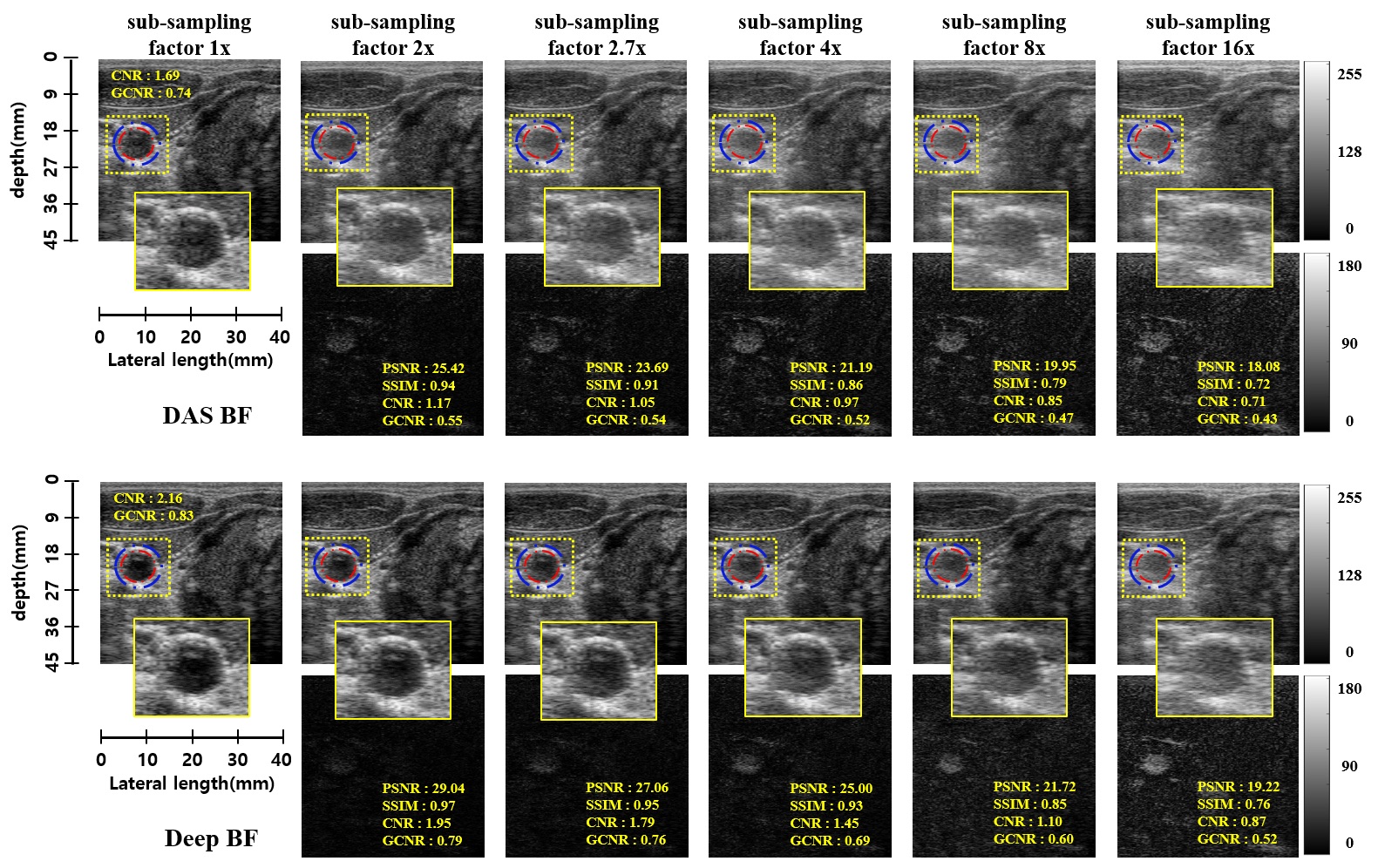, width=16cm}}   
	\centerline{\mbox{\footnotesize (a) Variable sampling scheme}}
	\vspace*{0.3cm}
	\centerline{\epsfig{figure=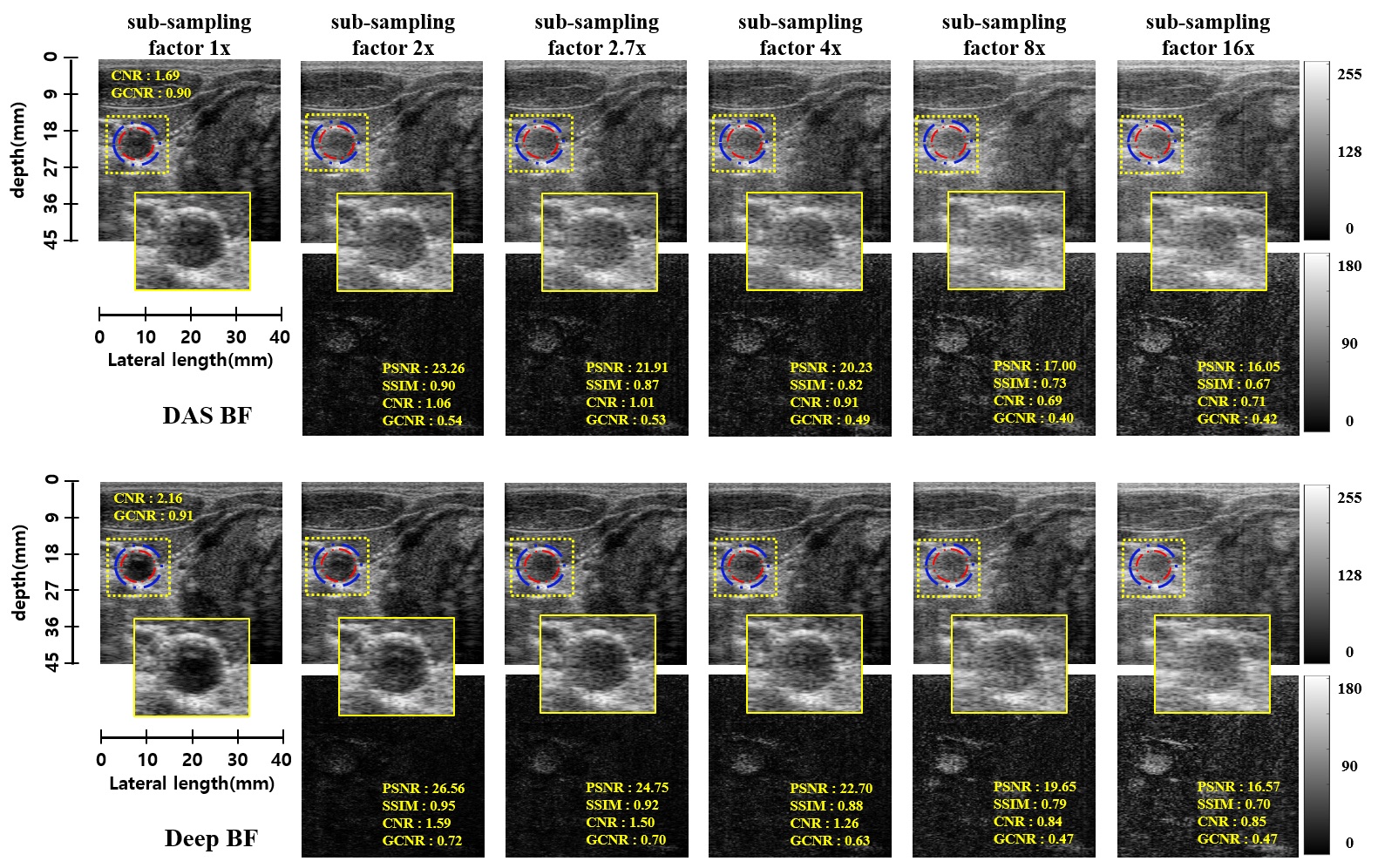, width=16cm}}   
	\centerline{\mbox{\footnotesize(b) Fixed sampling  scheme}}
%	  \centerline{\epsfig{figure=vdepth_invivo2.png, width=18cm}}   
	\caption{\footnotesize Reconstruction results of standard DAS beam-former and the proposed method for carotid region with respect to two subsampling scheme.}
		\label{fig:results_view_invivo1}	
\end{figure*} 

\begin{figure*}[!hbt]
	\centering
	\centerline{\epsfig{figure=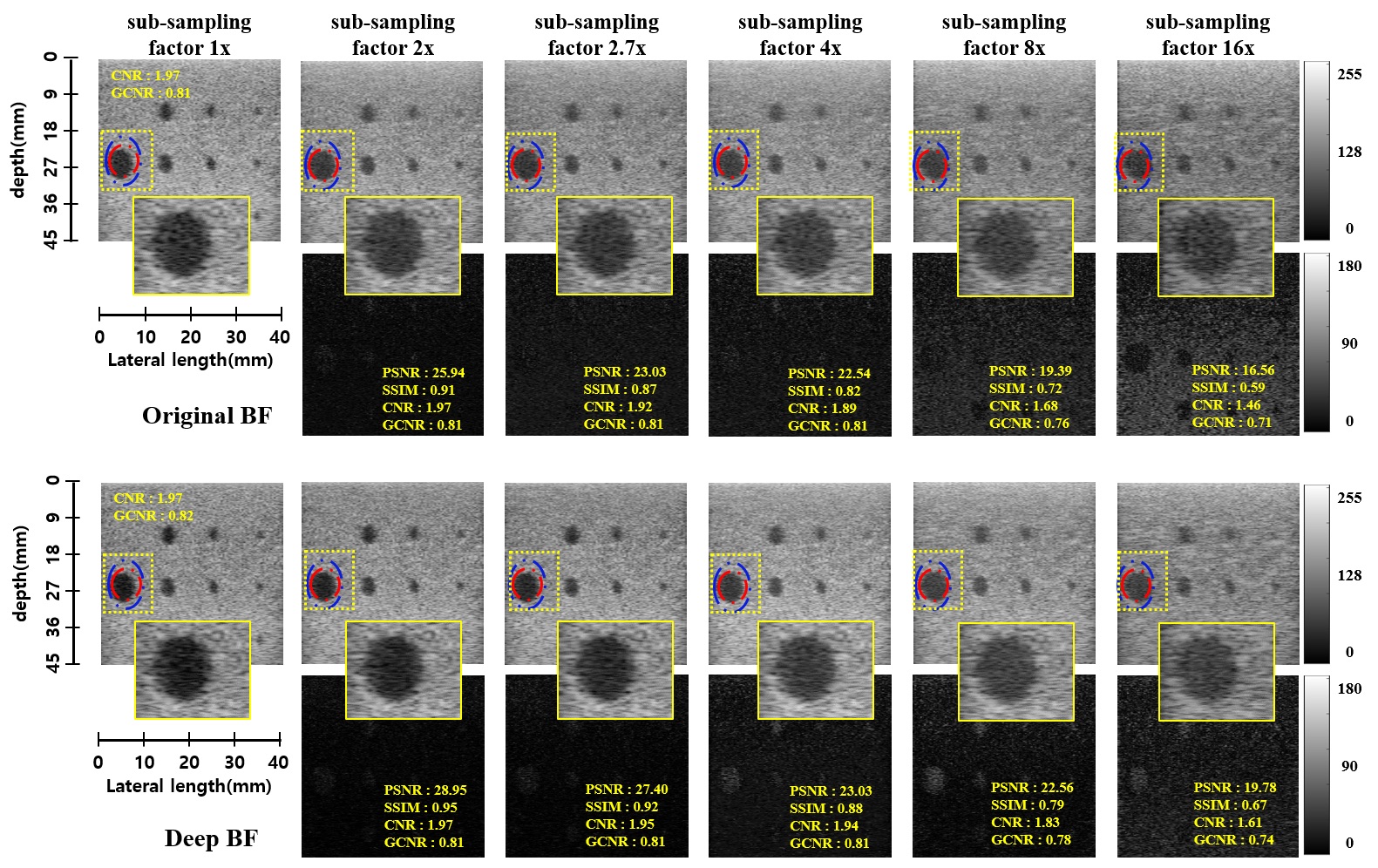, width=16cm}}   
	\centerline{\mbox{\footnotesize (a) Variable sampling scheme}}
		\vspace*{0.3cm}
	   \centerline{\epsfig{figure=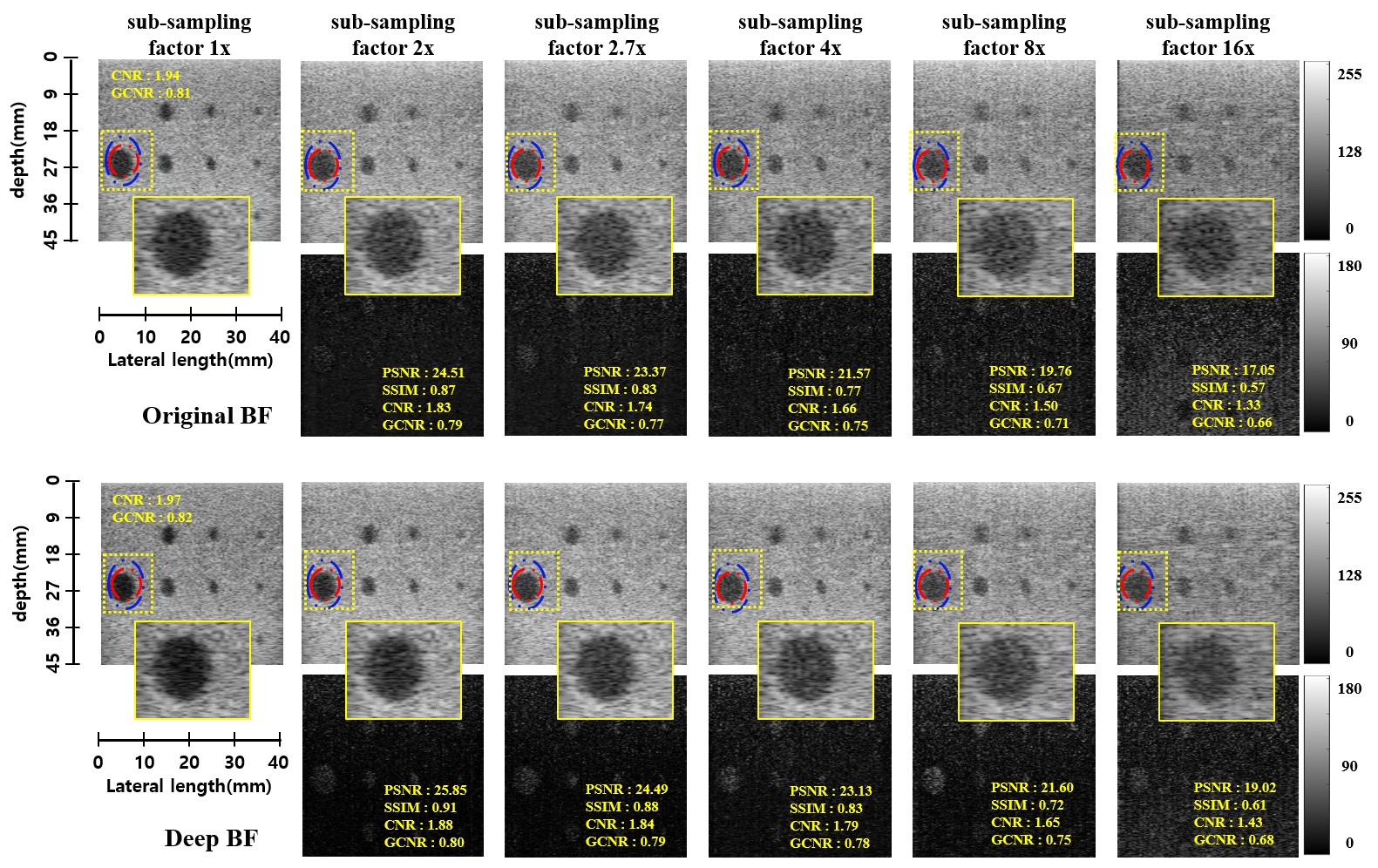, width=16cm}}   
	\centerline{\mbox{\footnotesize(b) Fixed sampling  scheme}}
	\caption{\footnotesize Reconstruction results of standard DAS beam-former and proposed method for phantom with respect to two subsampling scheme.  }
		\label{fig:results_view_phantom1}	
\end{figure*}

\section{Experimental Results}
\label{sec:results}

  Figs.~\ref{fig:results_view_invivo1}(a)(b)  show the results of two \textit{in vivo} examples for $64$, $32$, $24$, $16$, $8$ and $4$ Rx-channels down-sampling schemes using (a) variable sampling scheme  and (b) fixed sampling scheme.  Since 64 channels are used as a full sampled data,
   this corresponds to $1\times, 2\times, 2.7 \times, 4\times, 8\times$ and $16\times$ acceleration.  The images are generated using the proposed DeepBF and the  standard DAS beam-former method.   Our method significantly improves the visual quality of the US images by estimating the correct dynamic range and eliminating  artifacts for both sampling schemes.  From difference images in both figures, it is evident that under fixed down-sampling scheme, the quality degradation of images is higher than the variable sampling scheme, but the relative improvement in both schemes using the proposed method is nearly the same. Note that the proposed method successfully reconstruction both the near and the far field regions with equal efficacy, and  only minor structural details are imperceivable. 
 Furthermore, it is remarkable that the CNR and GCNR values are significantly improved by the DeepBF even for the fully sampled case (eg. from 1.69 to 2.16 in CNR and from 0.74 to 0.83 in GCNR for the case of
 variable sampling scheme), which clearly shows the
 advantages of the proposed method.

\begin{figure*}[!hbt]
	\centerline{\includegraphics[width=7cm]{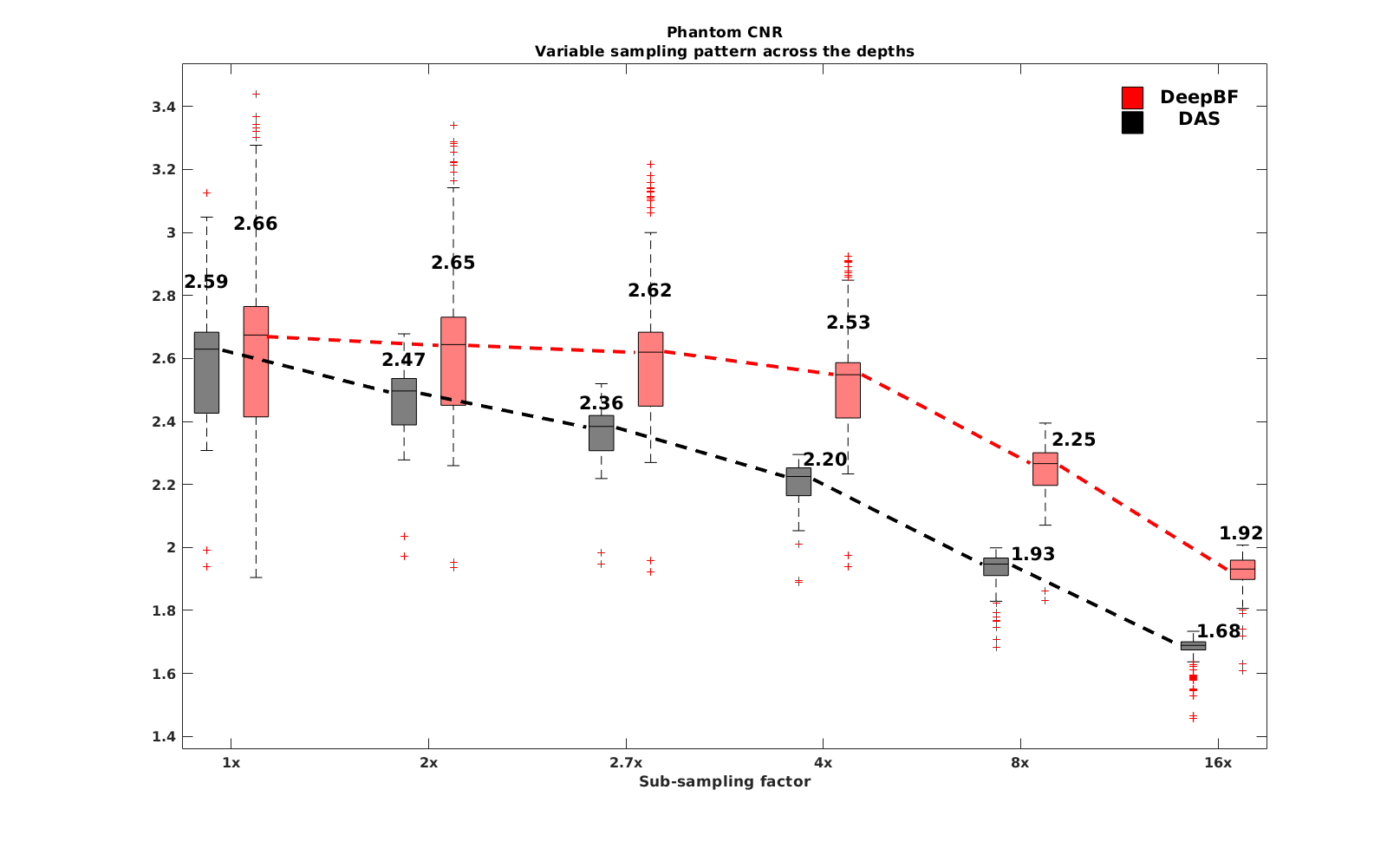}\hspace*{-0.5cm}\includegraphics[width=7cm]{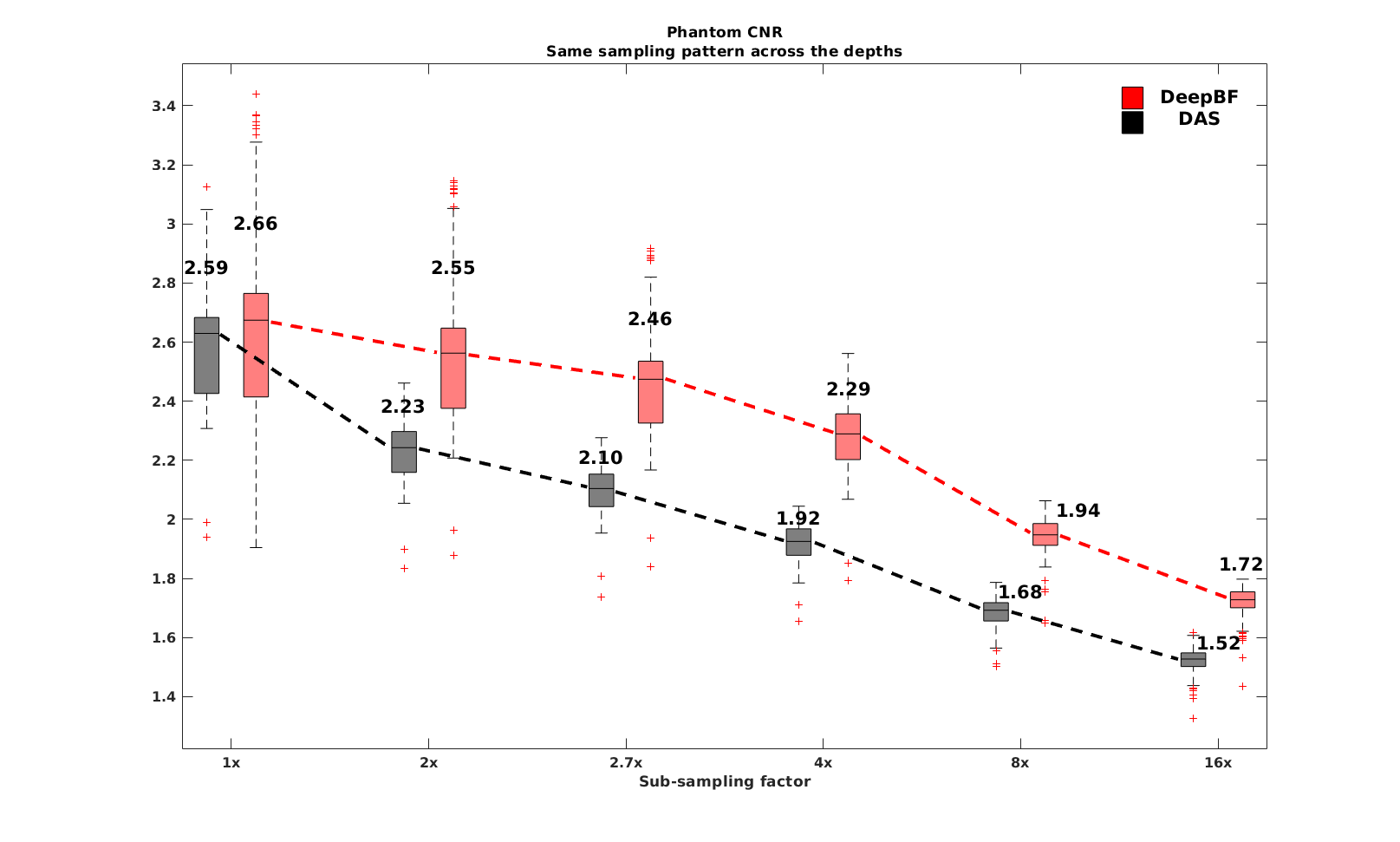}\hspace*{-0.5cm}}
	\vspace*{-0.3cm}
	\centerline{\mbox{\footnotesize(a) CNR value distribution }}
	\centerline{\includegraphics[width=7cm]{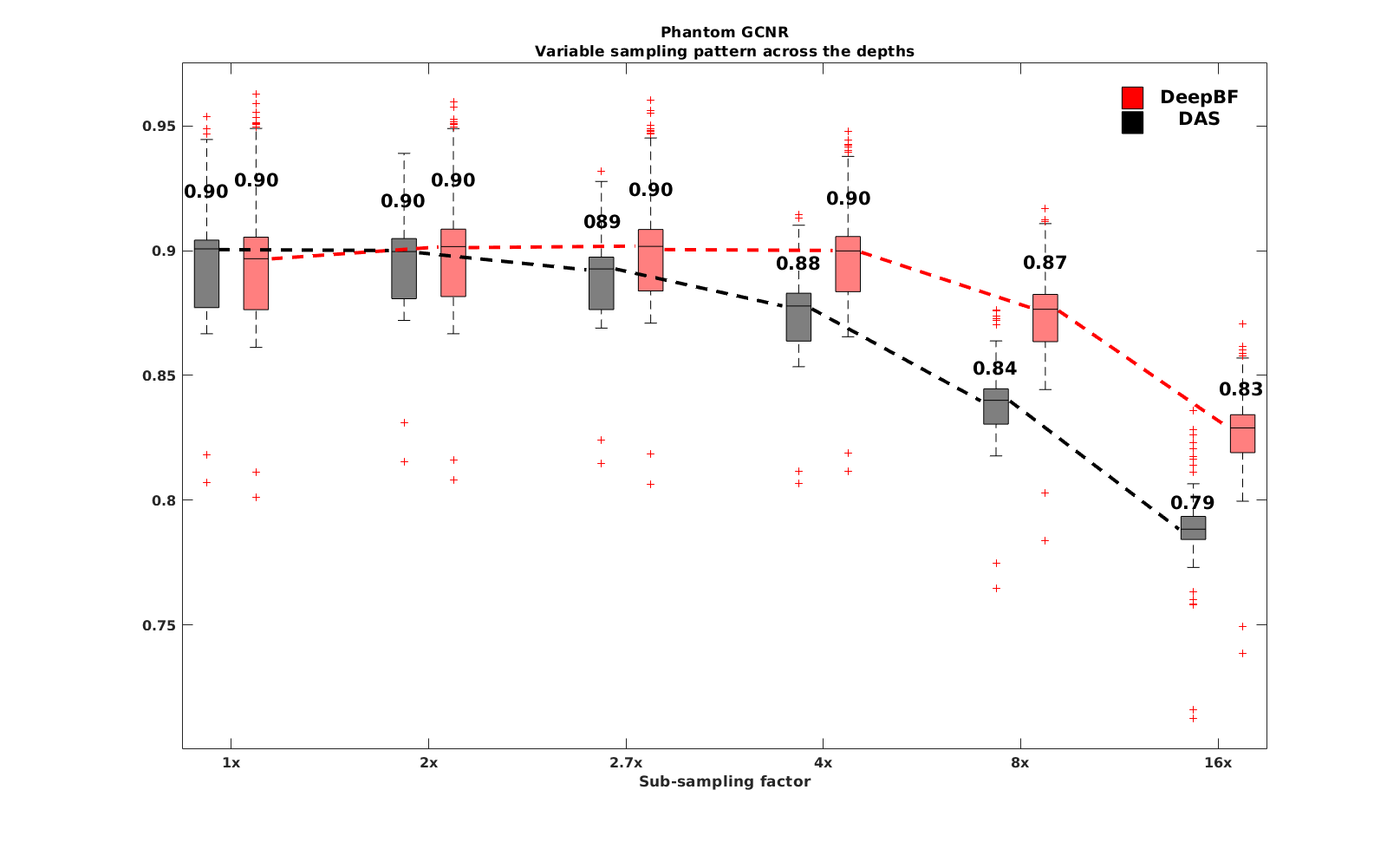}\hspace*{-0.5cm}\includegraphics[width=7cm]{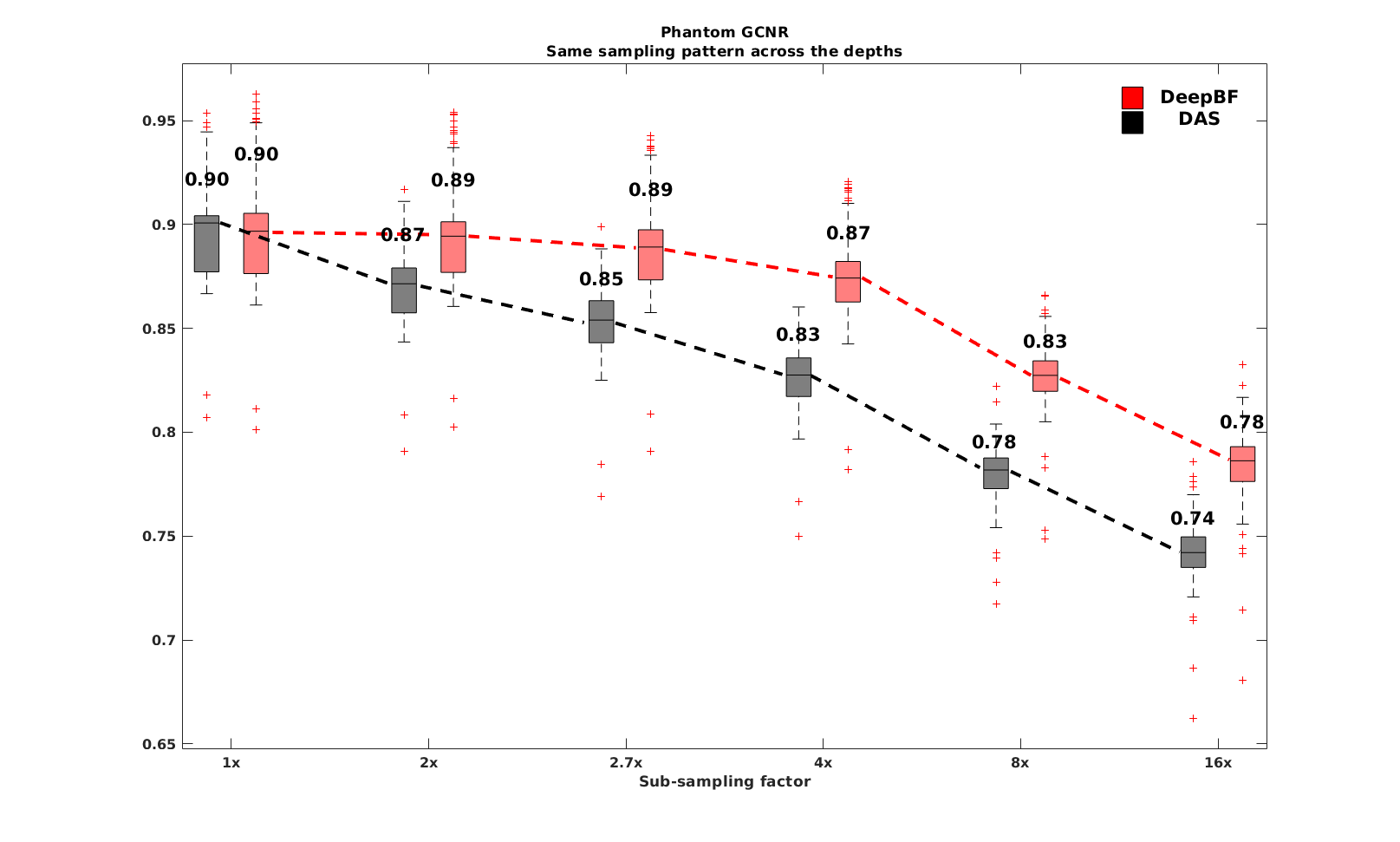}\hspace*{-0.5cm}}
		\vspace*{-0.3cm}
	\centerline{\mbox{\footnotesize(b) GCNR value distribution }}
%	\vspace*{0.3cm}
	\centerline{\includegraphics[width=7cm]{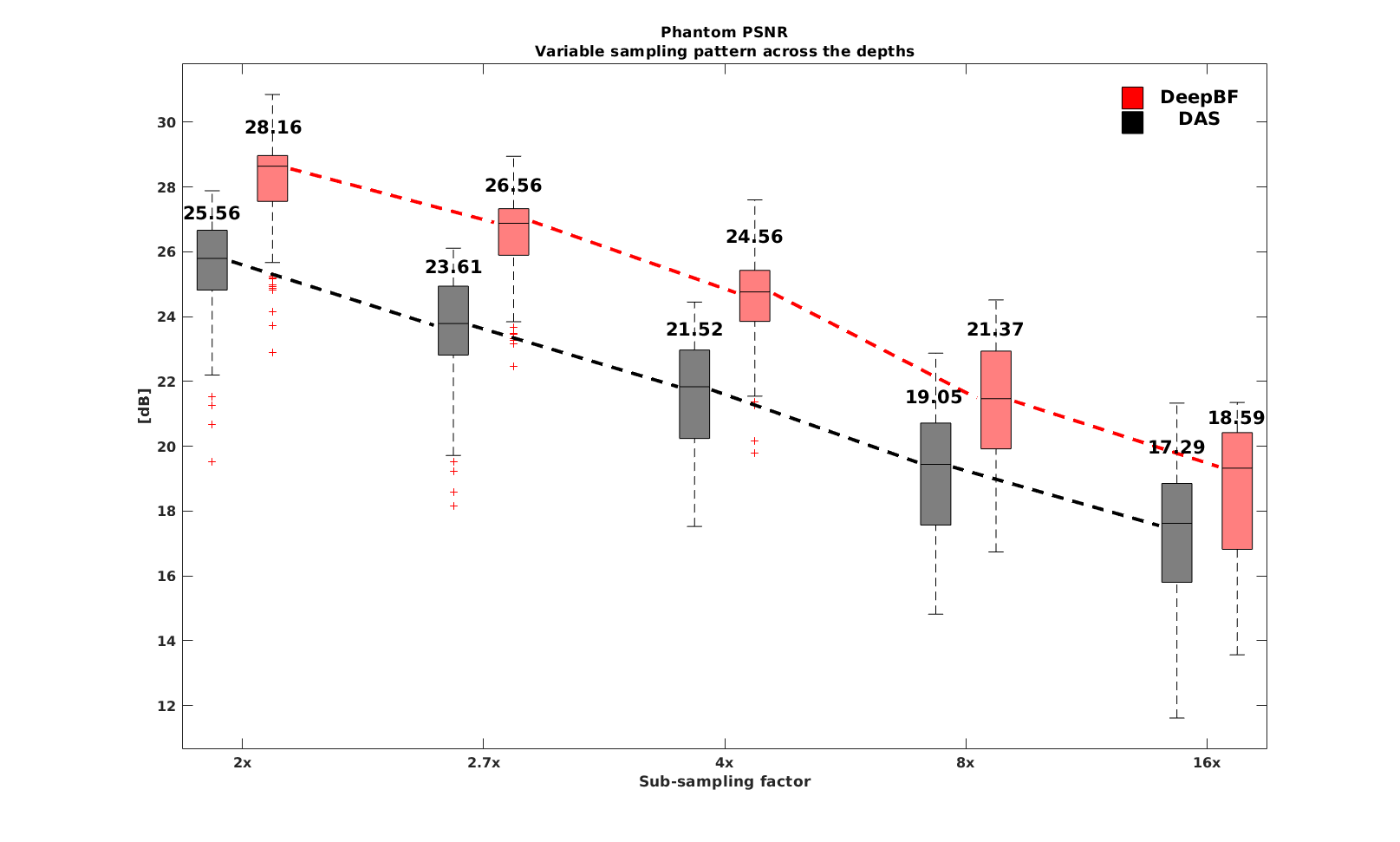}\hspace*{-0.5cm}\includegraphics[width=7cm]{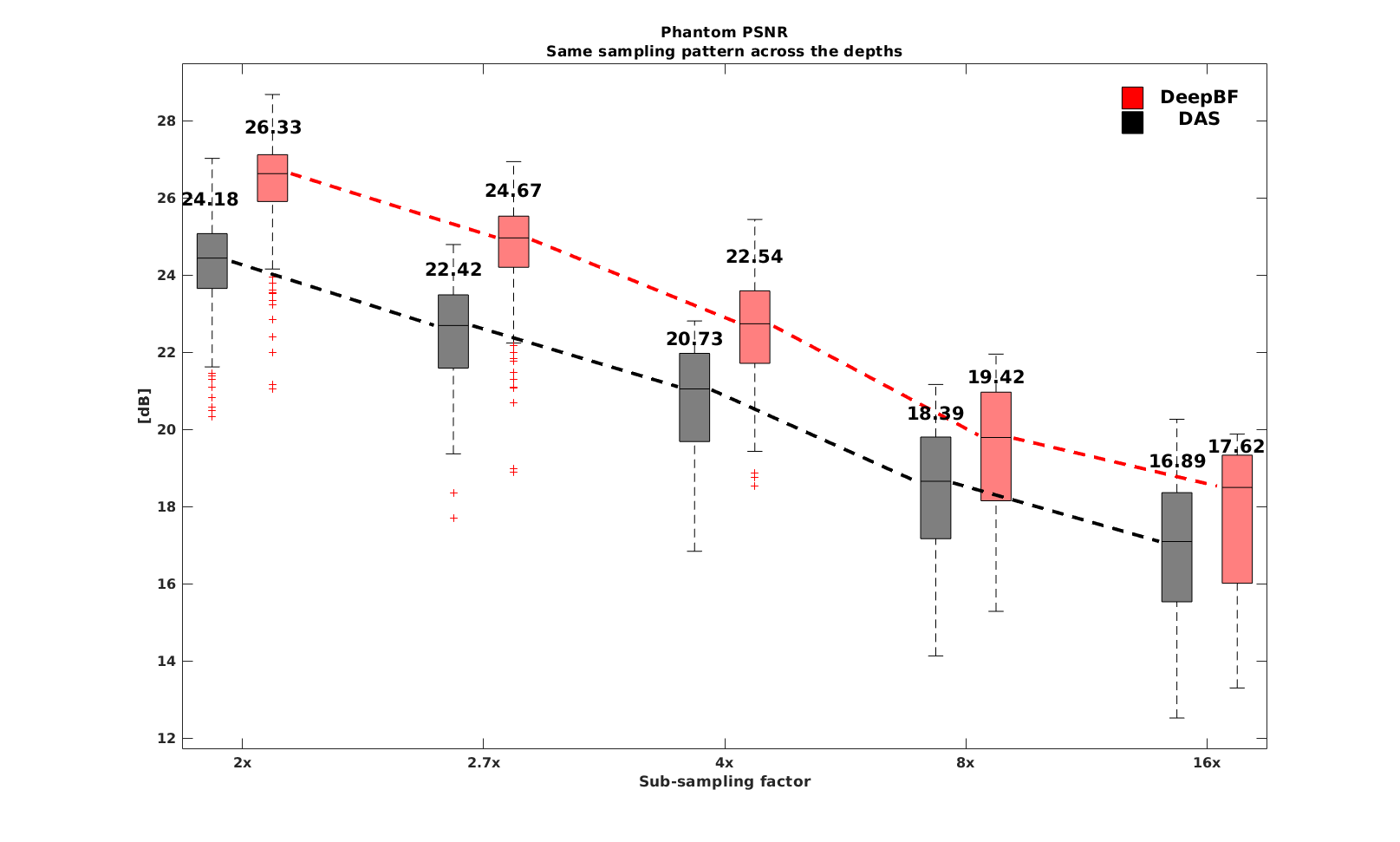}\hspace*{-0.5cm}}
		\vspace*{-0.3cm}
	\centerline{\mbox{\footnotesize(c) PSNR value distribution }}
%	\vspace*{0.3cm}
	\centerline{\includegraphics[width=7cm]{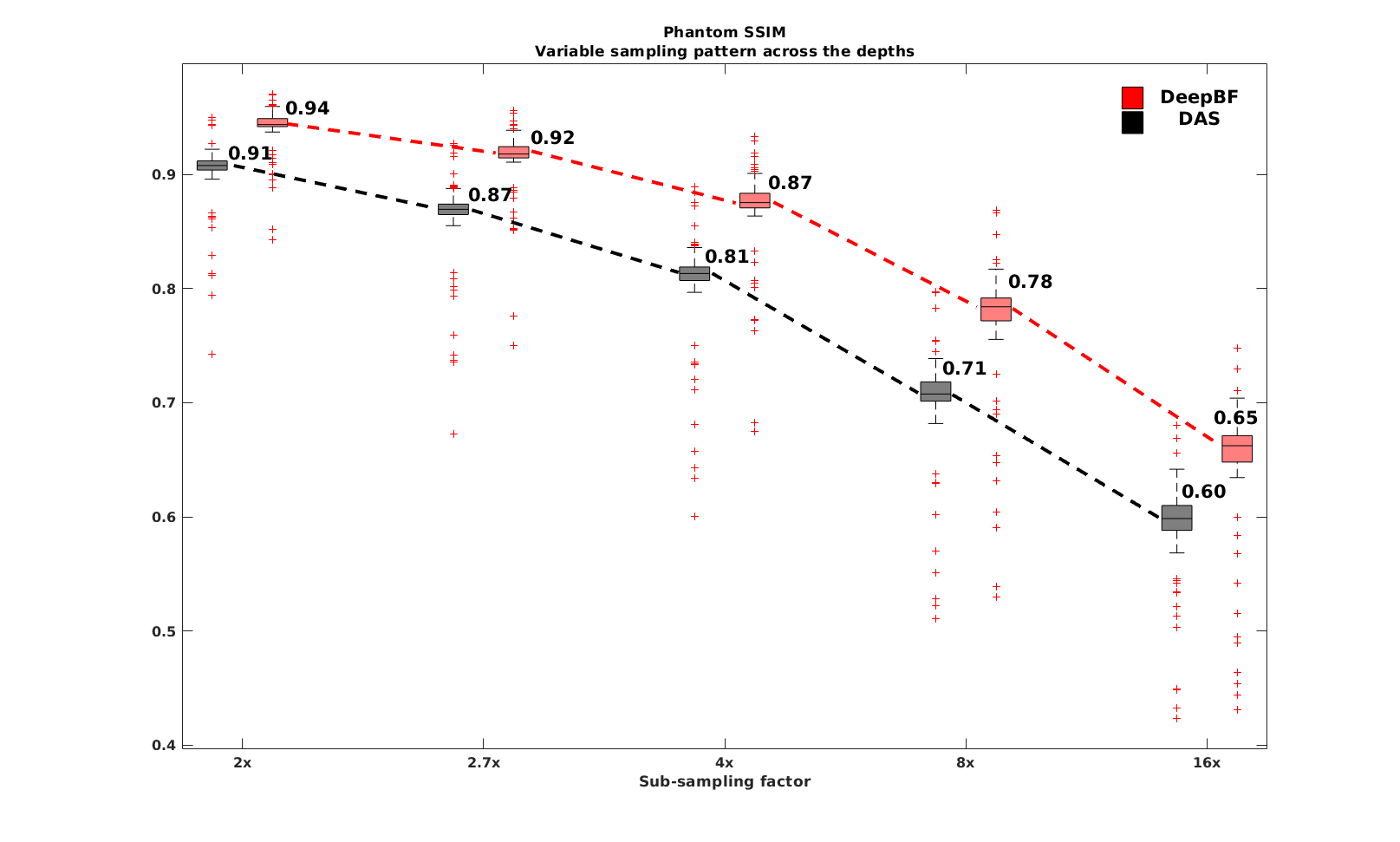}\hspace*{-0.5cm}\includegraphics[width=7cm]{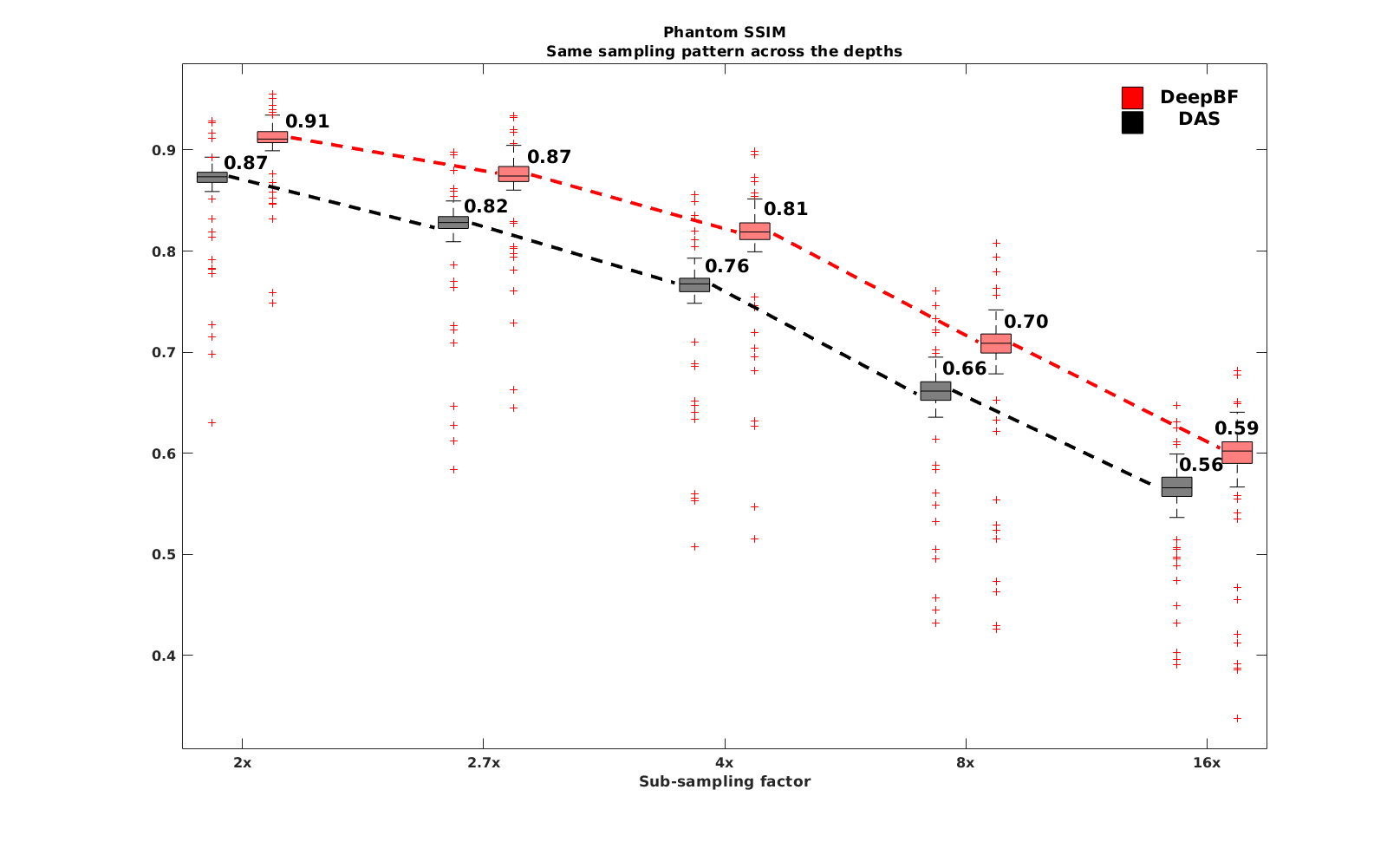}\hspace*{-0.5cm}}
		\vspace*{-0.3cm}
	\centerline{\mbox{\footnotesize(d) SSIM value distribution }}
%	\vspace*{0.3cm}
	\caption{\footnotesize Quantitative comparison using phantom data on different down-sampling schemes: ({first column}) variable sampling pattern cross the depth, ({second column})  fixed sampling pattern cross the depth}
	\label{fig:results_STATS_phantom}
\end{figure*}

%\begin{figure*}[!hbt]
%   \centering
%   \centerline{\epsfig{figure=sdepth_phantom1.png, width=18cm}}   
%	\vspace*{0.3cm}
%	\centerline{\epsfig{figure=sdepth_phantom2.png, width=18cm}}   
%	\vspace*{0.3cm}
%	\label{fig:results_view_phantom2}	
%	\caption{\bf\footnotesize Reconstruction results of standard DAS beam-former and proposed method. B-mode images of phantom from two sets of sub-sampled RF data.}
%\end{figure*}

   Fig.~\ref{fig:results_view_phantom1}(a)(b)  illustrate two representative examples of {phantom} data at 
   $1\times, 2\times, 2.7 \times, 4\times, 8\times$ and $16\times$ acceleration.
%   $64$, $32$, $24$, $16$, $8$ and $4$ Rx-channels. 
   By harnessing the spatio-temporal (multi-depth and multi-line) learning, the proposed CNN-based beam-former successfully reconstructs the images with good quality in all down-sampling schemes.  CNN automatically identifies the missing RF data and approximates it with available neighboring information.  Note that the network was trained on variable sampling scheme only;  however, the relative improvement in both schemes in test phase is nearly the same for both sampling
   schemes. This shows the generalization power of the proposed method.

\begin{table*}[!hbt]
	\centering
	\caption{Performance statistics on \textit{in vivo} data for variable sampling pattern}
	\label{tbl:results_vSTATS_invivo}
	\resizebox{0.5\textwidth}{!}{
	\begin{tabular}{c|cccccccc}
	\hline
		\multirow{2}{*}{DSR} & \multicolumn{2}{c}{\textbf{CNR}} & \multicolumn{2}{c}{\textbf{GCNR}} & \multicolumn{2}{c}{\textbf{PSNR (dB)}} & \multicolumn{2}{c}{\textbf{SSIM}}  \\
		& \textit{DAS} & \textit{DeepBF} & \textit{DAS} & \textit{DeepBF} & \textit{DAS} & \textit{DeepBF} & \textit{DAS} & \textit{DeepBF}  \\ \hline\hline
		1 & 1.38 & 1.45 & 0.64 & 0.66 & $\infty$ & $\infty$ & 1 & 1 \\
		2 & 1.33 & 1.47 & 0.63 & 0.66 & 24.59 & 27.38 & 0.89 & 0.95 \\
		2.7 & 1.3 & 1.44 & 0.62 & 0.66 & 23.15 & 25.54 & 0.86 & 0.92 \\
		4 & 0.25 & 1.38 & 0.6 & 0.64 & 21.68 & 23.55 & 0.81 & 0.87 \\
		8 & 1.18 & 1.26 & 0.58 & 0.6 & 19.99 & 21.03 & 0.74 & 0.77  \\
		16 & 1.12 & 1.17 & 0.56 & 0.58 & 18.64 & 19.22 & 0.67 & 0.69 \\  \hline
	\end{tabular}
}
\end{table*}

\begin{table*}[!hbt]
	\centering
	\caption{Performance statistics on \textit{in vivo} data for fixed sampling pattern}
	\label{tbl:results_sSTATS_invivo}
	\resizebox{0.6\textwidth}{!}{
		\begin{tabular}{c|cccccccc}
		\hline
			\multirow{2}{*}{Subsampling ratio} & \multicolumn{2}{c}{\textbf{CNR}} & \multicolumn{2}{c}{\textbf{GCNR}} & \multicolumn{2}{c}{\textbf{PSNR (dB)}} & \multicolumn{2}{c}{\textbf{SSIM}} \\
			& \textit{DAS} & \textit{DeepBF} & \textit{DAS} & \textit{DeepBF} & \textit{DAS} & \textit{DeepBF} & \textit{DAS} & \textit{DeepBF} \\ \hline\hline
			1 & 1.38 & 1.45 & 0.64 & 0.66 & 0 & 0 & 1 & 1 \\
			2 & 1.21 & 1.37 & 0.6 & 0.64 & 22.69 & 24.91 & 0.85 & 0.9 \\
			2.7 & 1.15 & 1.31 & 0.58 & 0.63 & 21.36 & 23.18 & 0.8 & 0.86 \\
			4 & 1.1 & 1.22 & 0.56 & 0.6 & 20.08 & 21.38 & 0.75 & 0.8 \\
			8 & 1.04 & 1.11 & 0.54 & 0.56 & 18.63 & 19.09 & 0.68 & 0.7 \\
			16 & 1.02 & 1.08 & 0.53 & 0.55 & 17.84 & 17.84 & 0.63 & 0.64 \\ \hline
		\end{tabular}
	}
\end{table*}

  We compared the CNR, GCNR, PSNR, and SSIM distributions of reconstructed B-mode images obtained from $188$ {phantom} test frames.  
In Fig.~\ref{fig:results_STATS_phantom}(a), for the variable sub-sampling scheme,  the proposed method achieved average CNR values of $2.66$, $2.65$, $2.62$, $2.53$, $2.25$, and $1.92$ in $64$, $32$, $24$, $16$, $8$ and $4$ Rx-channels down-sampling schemes, respectively, which are $2.70\%$, $7.29\%$, $11.02\%$, $15.00\%$, $16.58\%$ and $14.29\%$ higher than the standard DAS results. Whereas, on fixed sampling scheme, the proposed method achieved average CNR values of $2.66$, $2.55$, $2.46$, $2.29$, $1.94$, and $1.72$ in $64$, $32$, $24$, $16$, $8$ and $4$ Rx-channels down-sampling schemes, respectively. These values are $2.70\%$, $14.35\%$, $17.14\%$, $19.27\%$, $15.48\%$ and $13.16\%$ higher than the standard DAS results.
 
To test the robustness of our method we also evaluated the GCNR for all images. Fig.~\ref{fig:results_STATS_phantom}(b) compare the GCNR distributions for \textit{in vivo} and {phantom} data. Compared to standard DAS, the proposed deep beamformer  showed significant improvement for both sampling schemes at various subsampling factors.
Note that the GCNR of DeepBF images exhibit graceful degradation with respect to the subsampling factors in contrast to the DAS beamformer. 
 In fact, until $4\times$ subsampling, no performance degradation in GCNR was observed.  This again shows the robustness of the method.
 
 CNR, and GCNR, are the intensity differences measure for local regions, whereas the PSNR is the global intensity difference. 
 Fig.~\ref{fig:results_STATS_phantom}(c) compare the PSNR distributions. To calculate the PSNR, images generated from $64$ Rx-channels were considered as a reference image for all sampling schemes. Compared to standard DAS, using variable subsampling patterns,
% for \textit{in vivo} data using variable sampling scheme, the proposed deep learning method showed $2.79$dB, $2.39$dB, $1.87$dB, $1.04$dB, and $0.58$dB improvement on average for $32$, $24$, $16$, $8$ and $4$ Rx-channels down-sampling schemes, respectively. Whereas, for \textit{phantom} data, 
 the proposed deep learning method showed $2.6$dB, $2.95$dB, $3.04$dB, $2.32$dB, and $1.3$dB improvement on average for $32$, $24$, $16$, $8$ and $4$ Rx-channels down-sampling ratios (DSRs), respectively. Similar improvement was seen for the fixed downsampling scheme.
%, for \textit{in vivo} data on down-sampling scheme \# 2 , the proposed deep learning method showed $2.22$dB, $1.82$dB, $1.3$dB, $0.46$dB, and $0$dB improvement on average for $32$, $24$, $16$, $8$ and $4$ Rx-channels down-sampling schemes, respectively. 
Whereas, for fixed subsampling patterns, the proposed deep learning method showed $2.15$dB, $2.25$dB, $1.81$dB, $1.03$dB, and $0.73$dB improvement on average for $32$, $24$, $16$, $8$ and $4$ Rx-channels down-sampling schemes, respectively.

Another important measure of similarity is the structure similarity measure (SSIM). The higher SSIM means good recovery of detailed features of the image. To calculate the SSIM, images generated using $64$ Rx-channels were considered as reference images for all sampling schemes.  
As shown in  Fig.~\ref{fig:results_STATS_phantom}(d)
the proposed method shows significantly higher SSIM values for both sampling schemes.    
%Compared to standard DAS, for \textit{in-vivo} on down-sampling scheme \# 1, the proposed deep learning method showed $6$, $6$, $6$, $3$, and $2$ units improvement in $32$, $24$, $16$, $8$ and $4$ Rx-channels down-sampling schemes, respectively. Whereas, for \textit{phantom} data, the proposed deep learning method showed $3$, $5$, $6$, $7$, and $5$ units improvement in $32$, $24$, $16$, $8$ and $4$ Rx-channels down-sampling schemes, respectively. Similarly, for \textit{in-vivo} on down-sampling scheme \# 2, the proposed deep learning method showed $5$, $6$, $5$, $2$, and $1$ unit(s) improvement in $32$, $24$, $16$, $8$ and $4$ Rx-channels down-sampling schemes, respectively. Whereas, for \textit{phantom} data, the proposed deep learning method showed $4$, $5$, $5$, $4$, and $3$ units improvement in $32$, $24$, $16$, $8$ and $4$ Rx-channels down-sampling schemes, respectively. 
%The DeepBF produce images with SSIM value close to $1$ which means the structure of the reduced channel image is not very different from reference image.

  We also compared the CNR, GCNR, PSNR, and SSIM distributions of reconstructed B-mode images obtained from $360$ \textit{in-vivo}  test frames.
  Table~\ref{tbl:results_vSTATS_invivo} and ~\ref{tbl:results_sSTATS_invivo} showed that the proposed deep beamformer consistently outperformed the standard DAS beamformer for all
  subsampling scheme and ratio.  
%     In Fig.~\ref{fig:results_CNR}(a) for \textit{in-vivo} data, compared to DAS beamformer,  for the random sampling scheme,  the proposed method achieved average CNR values of $1.45$, $1.47$, $1.44$, $1.38$, $1.26$, and $1.17$ with $64$, $32$, $24$, $16$, $8$ and $4$ Rx-channels, respectively, which correspond to $5.07\%$, $10.52\%$, $10.77\%$, $10.4\%$, $6.78\%$ and $4.46\%$ improvement. For the fixed sampling scheme, the proposed method achieved average CNR values of $1.45$, $1.37$, $1.31$, $1.22$, $1.11$, and $1.08$ with $64$, $32$, $24$, $16$, $8$ and $4$ Rx-channel, respectively, whichare $5.07\%$, $13.22\%$, $13.91\%$, $10.91\%$, $6.73\%$ and $5.88\%$ improvement.

One big advantage of ultrasound image modality is it run-time imaging capability, which require fast reconstruction time. Another important advantage of the proposed method is the run-time complexity. Although training required $40$ hours for $200$ epochs using MATLAB, once training was completed, the reconstruction time for the proposed deep learning method is not very long. The average reconstruction time for each depth planes is around $9.8$ (milliseconds), which could be easily reduce by optimized implementation and reconstruction of multiple depth planes in parallel.

 \begin{figure*}[!hbt]
	\centerline{\includegraphics[width=16cm]{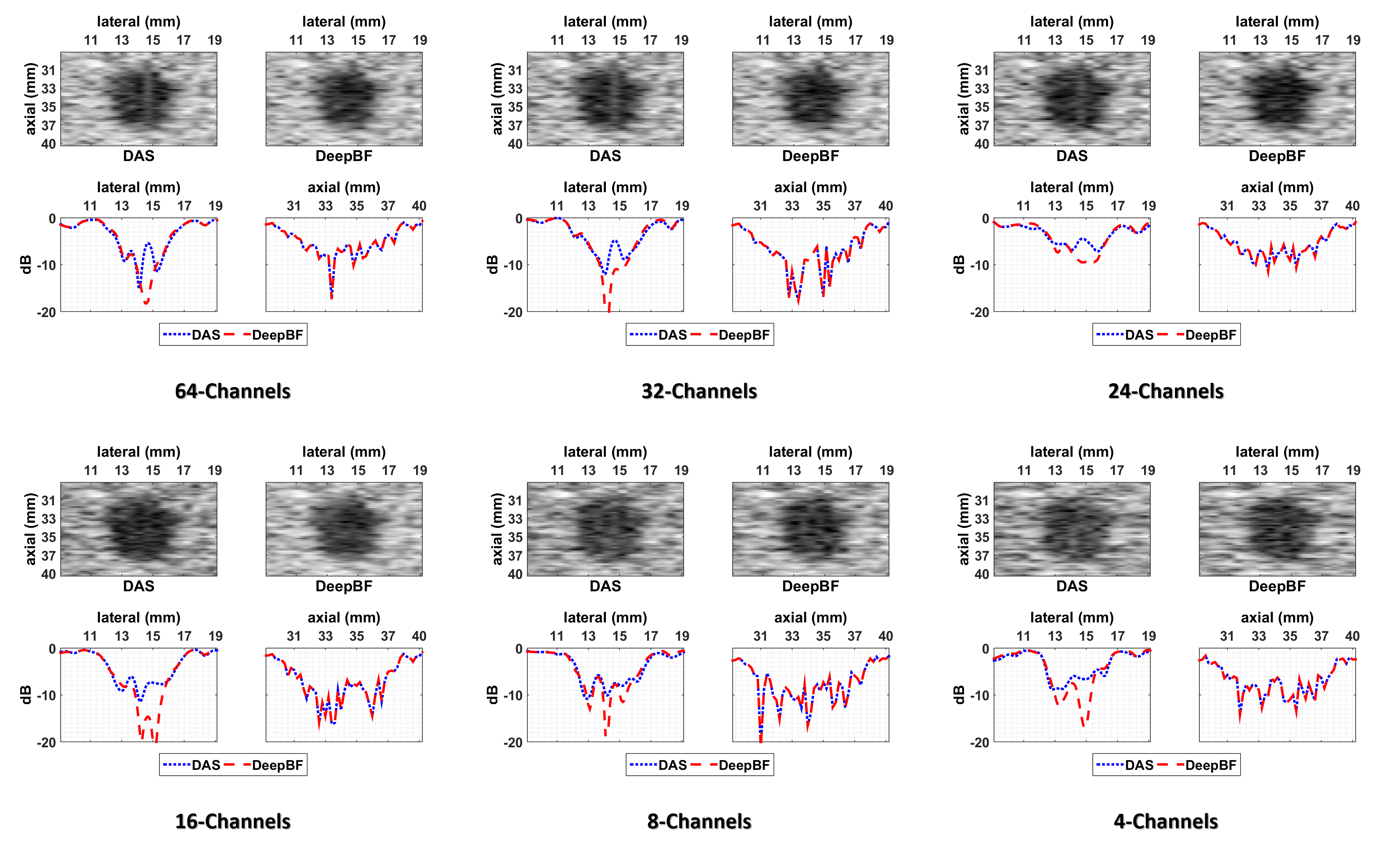}}
	\centerline{\mbox{\footnotesize(a) Phantom $6$ mm diameter anechoic cyst at $34$ mm from various RF sub-sampling rate.}}
	\vspace*{0.3cm}
	\centerline{\includegraphics[width=16cm]{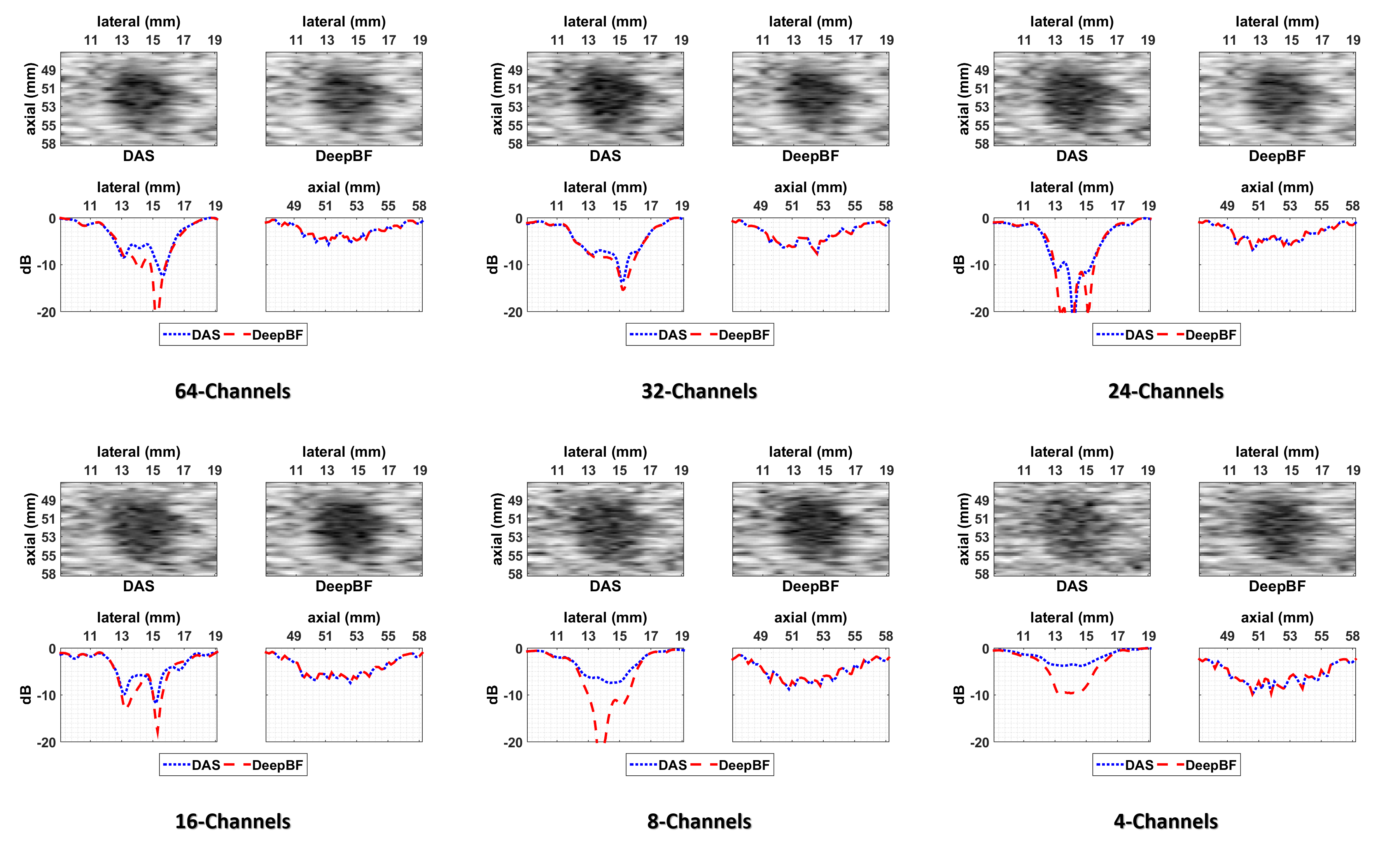}}	
	\centerline{\mbox{\footnotesize(b) Phantom $6$ mm diameter anechoic cyst at $52$ mm from various RF sub-sampling rate.}}
	\caption{Lateral and axial profiles through the center of the phantom anechoic cyst using DAS and DeepBF on random sampling across depth. Images are shown with a $60$ dB dynamic range.}
	\label{fig:results_SP_vLin}
\end{figure*}

\section{Discussion}
\label{sec:discussion}

We have designed a robust system which exploits the significant redundancies in the RF domains, which results in improved GCNR. It is noteworthy that  thanks to the exponentially increasing expressiveness of deep networks, for the first time a single {\em universal deep beamformer} trained using a purely data-driven way that can generate significantly improved images over widely varying aperture and channel sub-sampling patterns.  Moreover,  CNR, GCNR, PSNR, and SSIM were significantly improved over standard DAS method. Note that the proposed method efficaciously generate the better quality image from as little as only $6.25$\% RF-data.  

In Fig~\ref{fig:results_SP_vLin}, we compared lateral and axis profiles through the center of the two phantom anechoic cysts using DAS and DeepBF methods.  In particular, two anechoic cysts of $6$mm diameter scanned from the depth of $34$mm and $52$mm and B-mode images were obtained for random sampling scheme on $1\times$,$2\times$,$2.7\times$,$4\times$,$8\times$, and $16\times$ sub-sampling factors using DAS and our DeepBF. From the figures it can be seen that under all sampling schemes, on the boundary of cysts the proposed method show sharp changes in the pixel intensity with respect to the lateral position in the image. Although the axial profile shows similar trend to DAS, 
the average relative shift in the pixel intensity is constant for all sub-sampling factors, which means their is no significant degradation of axial resolution in sub-sampled images. 
The different resolution improvement between lateral and axial directions in the proposed method may be due to our training scheme in \eqref{eq:training}, which only consider
the three adjacent depth planes as input and average out the dependency with respect to $n$.  The depth dependent training scheme may be a solution for this, which will be investigated in other publications.

Note that our CNN is trained on full sampled ($64$-Rx) data, but surprisingly lateral resolution in DeepBF images is much better than the ($64$-Rx) images obtained from standard DAS method.  This \textit{super resolution effect} is prominent for both cysts obtained from different depths and the results are consistent cross the all sub-sampling factors.
This  is consistent with our observation on the
CNR and GCNR improvement on the full sampled data.
We believe that this is originated from the synergistic learning from multiple data set, which is not possible from analytic form of standard DAS beamformer.

\section{Conclusion}
\label{sec:conclusion}
In this paper, we presented a universal deep learning-based beamformer to generate high-quality B-mode ultrasound image from various rate of sub-sampled channel data. The proposed method is purely a data-driven method which exploits the spatio-temporal redundancies in the raw RF data, which help in generating improved quality B-mode images using fewer Rx channels. The proposed method improved the contrast of B-modes images by preserving the dynamic range and structural details of the RF signal in both the {phantom} and \textit{in-vivo} scans.
Due to the exponential large expressiveness of the deep neural network, our novel {\em universal} deep beamformer can efficiently  learn various mappings from RF measurements, and exhibits superior image quality for all sub-sampling rates.
Furthermore, the network was also used for  the fully sampled RF data  to significantly improve the image contrast and resolution. This {\em super-resolution} effects of neural network is shown in both {phantom} and \textit{in-vivo} images. Therefore, this method can be an important platform for RF sub-sampled US imaging.

%\bibliography{ref,refs_kh,convframelets,ref_icml}
%\bibliographystyle{naturemag}

\end{document}